\documentclass[12pt,letterpaper]{article}
\usepackage[a4paper, total={7in, 10in}]{geometry}

\usepackage{graphicx}
\usepackage{helvet}
\usepackage{authblk}
\usepackage{hyperref}
\usepackage{amsmath} 
\usepackage{amssymb} 
\usepackage{orcidlink} 
\usepackage[super,comma,sort&compress]  
   {natbib}

\usepackage{algorithm}
\usepackage{tcolorbox}
\usepackage{xcolor}
\usepackage{multirow}
\usepackage{tabularx}
\usepackage{booktabs} 
\usepackage{amsmath}
\usepackage{amssymb}
\usepackage{mathtools}
\usepackage{amsthm}
\usepackage{algorithm}
\usepackage{algorithmic}
\usepackage{makecell}
\usepackage{pifont}
\newcommand{\cmark}{\ding{51}} 
\newcommand{\xmark}{\ding{55}} 
\usepackage{adjustbox}
\usepackage[table]{xcolor}

\usepackage{comment}
\usepackage{listings}
\lstdefinestyle{jsonstyle}{
  basicstyle=\ttfamily\small,
  backgroundcolor=\color{gray!10},
  frame=single,
  breaklines=true,
  showstringspaces=false,
}

\makeatletter
\renewcommand{\maketitle}{\bgroup\setlength{\parindent}{0pt}
\begin{flushleft}
  \textbf{\@title}
  
  \@author
\end{flushleft}\egroup}
\makeatother

\title{\textcolor{black}{SelfCheck-Eval: A Multi-Module Framework for Zero-Resource Hallucination Detection in Large Language Models}}
\date{}

\author[1**\orcidlink{0009-0006-1178-4535}]{Diyana Muhammed}
\author[3,4** \orcidlink{0009-0008-3298-7168}]{Giusy Giulia Tuccari}
\author[2,*\orcidlink{0000-0002-1212-0101}]{Gollam Rabby}
\author[1\orcidlink{0000-0002-0698-2864}]{Sören Auer}
\author[1\orcidlink{0000-0002-7171-169X}]{Sahar Vahdati}

\affil[1]{TIB—Leibniz Information Centre for Science and Technology, Hannover, Germany}
\affil[2]{L3S Research Center, Leibniz University Hannover, Hannover, Germany}
\affil[3]{Department of Mathematics and Computer Science, University of Catania, Catania, Italy}
\affil[4]{National Research Council, Institute of Cognitive Science and Technology, Catania, Italy}

\affil[*]{Correspondence: gollam.rabby@l3s.de}
\affil[**]{These authors contributed equally}

\begin{document}

\maketitle

\section*{SUMMARY}
Large Language Models (LLMs) have demonstrated remarkable capabilities across diverse applications, from open-domain question answering to scientific writing, medical decision support, and legal analysis. However, their tendency to generate incorrect or fabricated content, commonly known as hallucinations, represents a critical barrier to reliable deployment in high-stakes domains.
Current hallucination detection benchmarks are limited in scope, focusing primarily on general-knowledge domains while neglecting specialised fields where accuracy is paramount. To address this gap, we introduce the AIME Math Hallucination dataset, the first comprehensive benchmark specifically designed for evaluating mathematical reasoning hallucinations. Additionally, we propose \textcolor{black}{SelfCheck-Eval}, a LLM-agnostic, black-box hallucination detection framework applicable to both open and closed-source LLMs.
Our approach leverages a novel multi-\textcolor{black}{module} architecture that integrates three independent detection strategies: the Semantic \textcolor{black}{module}, the Specialised Detection \textcolor{black}{module}, and the Contextual Consistency \textcolor{black}{module}.
Our evaluation reveals systematic performance disparities across domains: existing methods perform well on biographical content but struggle significantly with mathematical reasoning, a challenge that persists across NLI fine-tuning, preference learning, and process supervision approaches. These findings highlight the fundamental limitations of current detection methods in mathematical domains and underscore the critical need for specialised, black-box compatible approaches to ensure reliable LLM deployment.


\section*{KEYWORDS}


Hallucination Detection, Large Language Models, Consistency \& Uncertainty Quantification, Context‐based Verification

\section{Introduction}


Large Language Models (LLMs)~\citep{DBLP:conf/ijcai/WangSORRE24} have revolutionised natural language processing, excelling in tasks like summarisation, question answering, and dialogue generation. 
However, despite their impressive capabilities, LLMs can generate outputs that appear plausible but are factually incorrect, a phenomenon referred to as hallucination~\citep{DBLP:conf/emnlp/ManakulLG23}. 
Among the various types of hallucinations, input-conflicting, context-conflicting, and fact-conflicting, the latter poses the most significant challenges due to its potential to spread misinformation across critical domains like healthcare, finance, or education. 
Detecting and mitigating hallucinations is crucial to ensuring reliable LLMs in these sensitive areas~\citep{DBLP:journals/corr/abs-2410-02899}.

       

\begin{figure}[ht]
    \includegraphics[width=18cm]
    {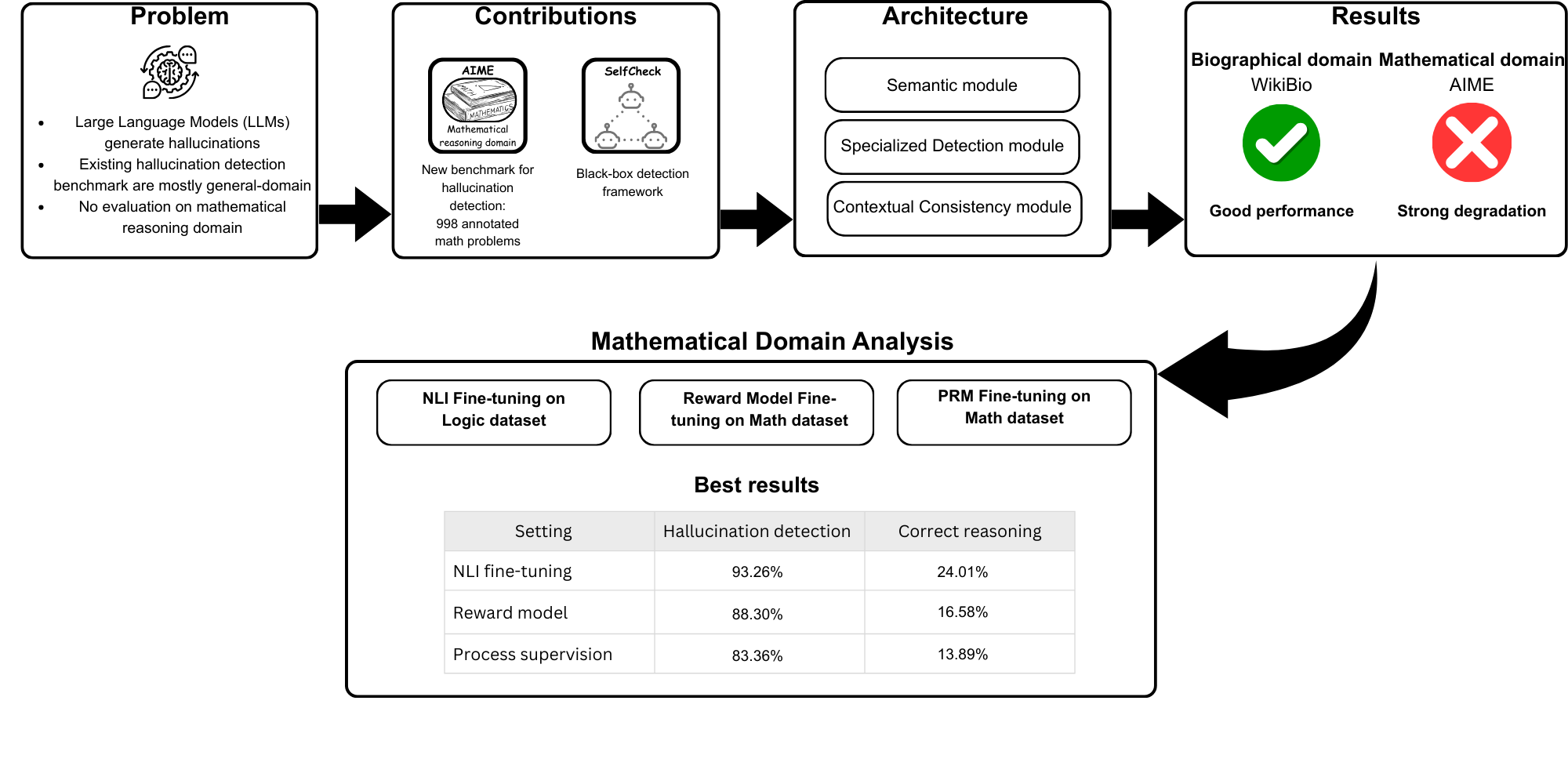}
    \caption{Overview of the hallucination detection experiment conducted within the \textcolor{black}{SelfCheck-Eval} framework.}
    \label{fig:use_example}
\end{figure}


\textcolor{black}{Hallucination detection methods must be evaluated across different domains to understand their generalisation capabilities. We focus on two complementary domains: biographical and mathematical reasoning. In biographical contexts, hallucinations appear as incorrect facts, while in mathematical reasoning, hallucinations involve flawed logic in multi-step problem solving. This tests whether detection methods work across content types or domain-specific approaches, a critical question for reliable LLM deployment.}

Traditional hallucination detection approaches~\cite{DBLP:conf/emnlp/0038GLZPK24} focus primarily on static claim-evidence pairs and cannot handle the dynamic context intrinsic to LLM-generated content.
Moreover, existing benchmarks for hallucination detection~\cite{hallucinations-leaderboard} mainly assess basic factual consistency in general domains, often neglecting the unique challenges posed by complex mathematical reasoning where both logical consistency and factual correctness are required. 
This narrow scope leaves critical gaps in understanding how different detection paradigms perform across domains with varying reasoning requirements, particularly in mathematical contexts where current methods may face fundamental limitations.
\textcolor{black}{As shown in Table~\ref{tab:comparison}, existing datasets focus primarily on general domains, and they lack sampled LLM responses, which is a major factor for our evaluation approach.}

\begin{table*}[htbp]
\centering
\resizebox{\textwidth}{!}{%
\scriptsize
\begin{tabular}{llllccc}
\hline
\textbf{Dataset} & \textbf{Source} & \textbf{Domain} & \textbf{Annotation} & \textbf{Logical reasoning} & \textbf{LLM-Sampled} & \textbf{Usefulness} \\ 
\hline
\textbf{TruthfulQA}~\cite{DBLP:conf/acl/LinHE22} & General & Text & Manual & \xmark & \xmark & \xmark \\ 
\hline
\textbf{HaluEval}~\cite{DBLP:conf/emnlp/LiCZNW23} & Text & General & Manual & \xmark & \xmark & \xmark \\ 
\hline
\textbf{HADES}~\cite{DBLP:conf/acl/LiuZBMSCD22} & Token level & General & Crowdsourced & \xmark & \xmark & \xmark \\ 
\hline
\textbf{FACTCHD}~\cite{DBLP:conf/ijcai/0016SGW000LZC24} & KGs \& Text & General & Manual & \xmark & \xmark & \xmark \\ 
\hline
\textbf{WikiBio}~\cite{DBLP:conf/acl/StranisciDMPRC23} & Text & Biography & Manual & \xmark & \cmark & \cmark \\ 
\hline
\textbf{AIME (Ours)} & Math Problems & Mathematics & AI-Human & \cmark & \cmark & \cmark \\ 
\hline
\end{tabular}%
}
\caption{Comparison with existing hallucination detection datasets. Our AIME includes complex mathematical domains, such as number theory, geometry, algebra, and others.}
\label{tab:comparison}
\end{table*}

To address these gaps, we introduce \textcolor{black}{SelfCheck-Eval}, a systematic comparative \textcolor{black}{framework} that evaluates three distinct detection \textcolor{black}{modules} across biographical and mathematical domains. 

\textcolor{black}{Our methods are designed for zero-resource black-box hallucination detection. The black-box constraint means we operate without access to LLM internals. The zero-resource constraint means we do not use external knowledge bases or fact-checking systems. Instead, our methods rely analysing analyzing multiple text outputs generated by the LLM being evaluated. This design makes them applicable to any LLM, including closed-source systems.}

Our framework examines: (1) statistical approaches based on frequency analysis and response consistency, (2) neural methods \textcolor{black}{using} fine-tuned LLMs for entailment detection, and (3) contextual approaches \textcolor{black}{using} LLM reasoning capabilities through structured prompting. 
Rather than proposing a unified detection \textcolor{black}{system}, we provide empirical insights into how different \textcolor{black}{methods} perform when applied to domains with varying complexity and reasoning requirements.
In \autoref{fig:use_example}, each detection approach independently evaluates LLM responses and provides hallucination scores.

\begin{figure}
	\centering
		\includegraphics[scale=.4]
        {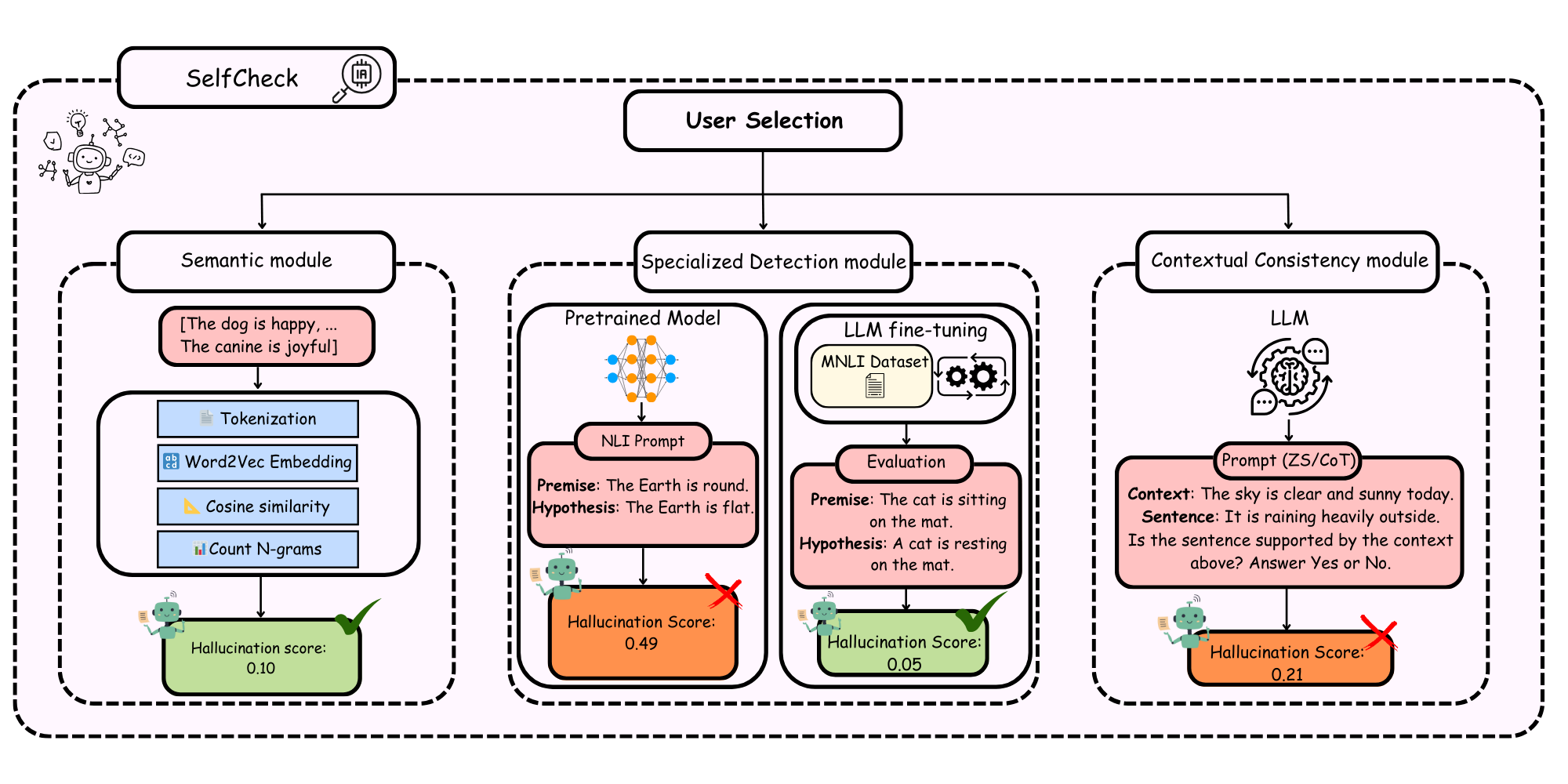}
	\caption{Overview of the \textcolor{black}{SelfCheck-Eval} pipeline with the approaches for hallucination detection: Semantic \textcolor{black}{module}, Specialized Detection \textcolor{black}{module} (pre-trained and fine-tuned), and Contextual Consistency \textcolor{black}{module}.}
	\label{fig:pipeline}
\end{figure}

\textcolor{black}{Our investigation proceeds in two phases. First, we systematically evaluate three detection paradigms across biographical (WikiBio) and mathematical 
(AIME) domains (Section ~\ref{main_results}). Second, observing consistent failures in mathematical 
reasoning validation, we investigate whether specialized fine-tuning on mathematical datasets can overcome these limitations through NLI fine-tuning, reward modeling, and process supervision approaches (Section~\ref{analysis-math}).
To enable this systematic evaluation, we introduce the AIME Math Hallucination benchmark. Unlike previous approaches that rely on artificially injected hallucinations, our benchmark uses mathematical reasoning errors naturally generated by state-of-the-art LLMs, validated through human-AI collaboration. This provides a realistic testbed for assessing detection methods on authentic mathematical reasoning challenges.
Our evaluation across the WikiBio and AIME datasets reveals significant performance variations between domains. While detection methods achieve strong performance on biographical content, they exhibit consistently asymmetric behavior in mathematical reasoning. Methods effectively detect incorrect solutions but systematically struggle to validate correct reasoning. This pattern emerges across different approaches, suggesting fundamental challenges in current methods when applied to mathematical domains.}\textcolor{black}{The main contributions of this work are:
\begin{itemize}
\item We provide a systematic comparative analysis of black-box zero-resource hallucination detection paradigms across domains, revealing significant 
performance variations between biographical and mathematical content.
\item We introduce the AIME Math Hallucination benchmark featuring naturally 
occurring mathematical reasoning errors generated by state-of-the-art LLMs, 
providing a realistic evaluation testbed beyond existing datasets that rely 
on artificial hallucination injection.
\item We observe consistent asymmetric performance patterns across multiple 
detection paradigms in mathematical reasoning, with high sensitivity to 
errors but systematic difficulty validating correct solutions.
\item We evaluate specialized fine-tuning approaches, including NLI-based fine-tuning, reward modeling, and process supervision, finding that the mathematical reasoning challenge continues across training paradigms.
\end{itemize}
}

\section{Related Work}

LLMs demonstrate remarkable abilities in understanding and executing user instructions~\cite{DBLP:conf/emnlp/Peng0S24}. However, they frequently generate misleading outputs, termed \textrm{``hallucinations''}~\cite{DBLP:conf/www/ShenLFRBN23}. These hallucinations are categorized into input-conflicting, context-conflicting, and fact-conflicting types, with fact-conflicting hallucinations posing the most significant challenges due to their potential to propagate misinformation~\cite{DBLP:conf/ijcai/0016SGW000LZC24}.

Previous research has investigated hallucinations across various NLP tasks, including summarization~\cite{DBLP:conf/naacl/PagnoniBT21} and dialogue generation~\cite{DBLP:conf/emnlp/DasSS22}, alongside other applications~\cite{DBLP:conf/acl/MallenAZDKH23}. Self-consistency decoding has shown promise in improving reasoning accuracy by leveraging multiple reasoning paths~\cite{rabby2024mc}, yet its effectiveness in detecting factual inaccuracies remains underexplored. Despite these advances, current methodologies exhibit critical limitations: lack of standardized evaluation metrics for fact-conflicting hallucinations, absence of domain-specific benchmarks for mathematical reasoning, and insufficient focus on black-box detection methods suitable for API accessible LLMs. Frequency-based methods rely on statistical patterns and token probability estimation from multiple LLM outputs, such as n-gram analysis and distributional consistency measures. Neural approaches leverage fine-tuned LLMs for natural language inference (NLI) or entailment detection, often requiring supervised fine-tuning on domain-specific data. Contextual methods exploit the reasoning capabilities of LLMs themselves through carefully designed prompts and consistency evaluation. While each paradigm has demonstrated effectiveness in specific scenarios, comprehensive comparative analysis across different domains, particularly in mathematical reasoning, remains underexplored. Perturbation-based approaches~\cite{DBLP:conf/naacl/ThorneVCM18} and synthetic data generation methods~\cite{DBLP:conf/acl/LiuZBMSCD22} artificially modify factual texts to create hallucinatory data, potentially missing the natural variation patterns of real LLM hallucinations. 
White-box methods~\cite{DBLP:conf/ijcai/0016SGW000LZC24} require supervised data and access to internal LLM states, while self-evaluation approaches~\cite{DBLP:conf/acl/ZhangPTZJSMM24} show limited generalizability across different tasks and domains.

Mathematical reasoning presents unique challenges for hallucination detection that are not adequately addressed by existing approaches designed for general domain text. Mathematical content requires both logical consistency across multi-step derivations and factual correctness of individual claims, making it qualitatively different from biographical or narrative text, where surface-level inconsistencies may suffice for detection. Existing works in mathematical reasoning typically employ standard math benchmarks like GSM8K \cite{gsm8k} or MATH \cite{math} and create hallucinated data through artificial injection methods, rather than analyzing naturally occurring mathematical reasoning errors.

Recent work introduces FG-PRM~\cite{fgprm2024}, which proposes a fine-grained taxonomy of six hallucination types for mathematical reasoning tasks. Their approach focuses on step-level detection using specialized Process Reward Models (PRMs) \cite{lightman2023letsverify} trained on synthetically generated hallucination data and validated on human-annotated solutions. However, their synthetic data generation involves systematically injecting predefined hallucination types into correct solutions, which may not capture the authentic distribution of naturally occurring mathematical reasoning errors that emerge during actual LLM deployment.

Farquhar et al.~\cite{farquhar2024} introduce semantic entropy as a statistical framework for hallucination detection. The method estimates uncertainty by sampling multiple responses, clustering them according to their semantic similarity via bidirectional entailment, and computing entropy over the cluster distribution. While this approach performs well for open domain question answering, its evaluation on mathematical reasoning was limited to SVAMP \cite{svamp}, a dataset of elementary grade school word problems. The application of more complex mathematical reasoning presents unique challenges. In mathematical contexts, solutions often involve long reasoning chains followed by numerical answers. Two solutions may appear semantically close because they share intermediate reasoning, yet still diverge in the final step, leading to different outcomes. Conversely, reasoning chains with different surface forms can nonetheless converge to the same correct answer. Thus, when applying semantic entropy to mathematical reasoning, both the coherence of the reasoning process and the correctness of the final answer need to be considered.

The effectiveness of different detection paradigms across mathematical versus general domain content remains underexplored, with most evaluations focusing on single-domain performance rather than systematic cross-domain analysis. Our work addresses this gap by introducing a systematic comparative analysis framework that evaluates three distinct detection paradigms across both biographical and mathematical domains. Rather than proposing a unified detection method, we provide empirical insights into the relative strengths and limitations of different approaches when applied to domains with varying complexity and reasoning requirements.

\begin{table}[ht]
\centering
    \resizebox{\textwidth}{!}{%
    \scriptsize
    \begin{tabular}{p{2cm}|p{3cm}|p{3cm}|p{2cm}|p{2cm}}
    \hline
    \textbf{Dataset} & \textbf{Major Inaccurate} & \textbf{Minor Inaccurate} & \textbf{Accurate} & \textbf{Total} \\ \hline
    WikiBio & 761 & 516 & 631 & 1,908 \\ \hline
    AIME    & 823 & 38  & 137 & 998   \\ \hline
    \end{tabular}%
    }
\caption{Sample distribution of the experimental datasets.}
\label{tab:sample_distribution}
\end{table}

\section{Datasets and Metrics}
\label{sec:datasets}
In this section, we present the datasets employed in our comprehensive analysis for detecting hallucinations across different domains. Our evaluation framework encompasses two complementary benchmarks that together provide a robust assessment of hallucination detection capabilities.

We begin by examining WikiBio~\cite{DBLP:conf/emnlp/ManakulLG23}, an established general knowledge domain benchmark for hallucination detection, which serves as our baseline for comparative analysis. Subsequently, we introduce AIME-MATH Hallucination, a novel benchmark specifically designed for evaluating and detecting hallucinations in mathematical reasoning tasks. This new benchmark addresses the critical gap in domain-specific evaluation tools for mathematical domains. Finally, we provide a detailed description of the AIME-MATH construction methodology, highlighting the design principles and validation procedures that ensure its effectiveness for mathematical hallucination detection. The sample distributions of these datasets are summarized in Table~\ref{tab:sample_distribution}.

\subsection{WikiBio Dataset}

The WikiBio dataset represents a well-established benchmark for hallucination detection in general knowledge domains, comprising 238 Wikipedia paragraphs paired with corresponding tabular data. The dataset construction relied on synthetic passages generated with GPT-3.5 \cite{DBLP:conf/emnlp/ManakulLG23},through a standardized prompt template: \textit{``This is a Wikipedia passage about [concept]:''}.\\
This approach ensures consistent generation conditions while covering diverse biographical topics.
The resulting dataset contains 1,908 sentences with an average length of 184.7 tokens per sentence. Each sentence underwent comprehensive factuality annotation using a three-level classification system: Major Inaccurate (39.9\%), Minor Inaccurate (33.1\%), and Accurate (27.0\%). The annotation process demonstrates robust inter-annotator reliability, with Cohen's $\kappa$ scores of 0.595 for the full three-class distinction and 0.748 for binary accurate/inaccurate classification. For quality assurance, 201 sentences received dual annotation, with disagreements resolved using a conservative worst-case labeling approach.
The dataset's design enables both sentence-level and passage-level evaluation. Passage-level hallucination scores, computed by averaging constituent sentence scores, range up to +1.0, indicating complete factual inaccuracy. This multi-granular annotation framework provides a comprehensive foundation for evaluating hallucination detection methods across different levels of textual analysis.

\subsection{AIME Math Hallucination Benchmark}
The AIME-MATH Hallucination dataset addresses the critical need for domain-specific benchmarks in mathematical reasoning by leveraging problems from the American Invitational Mathematics Examination (AIME). Our construction methodology follows a systematic three-phase approach designed to ensure both scalability and annotation quality.
\subsubsection{Data Collection and Response Generation}
The initial phase involved a comprehensive collection of human solutions for all 998 AIME problems, establishing verified ground truth solutions with detailed explanations for subsequent evaluation. To capture the variability inherent in LLM mathematical reasoning, we generated five distinct responses per problem using GPT-4o with zero-shot prompting. This multi-response approach enables robust evaluation of consistency patterns and provides sufficient sample diversity for statistical analysis of hallucination detection methods.
\subsubsection{Human-AI collaboration Annotation}

\textcolor{black}{
Our annotation methodology combines automated processing with human oversight to achieve both efficiency and accuracy. For each problem, we randomly selected one response from the five generated samples and evaluated its correctness against the human-provided ground truth solution. The initial categorization employed a binary matching criterion: responses producing answers consistent with the ground truth were preliminarily classified as potentially accurate, while those with incorrect final answers were automatically designated as Major Inaccurate.}

\textcolor{black}{To refine this initial classification, we implemented a hybrid human-AI collaboration framework utilizing GPT-3.5 for systematic answer matching, supplemented by human annotator review for ambiguous cases. This approach leverages the efficiency of automated processing while maintaining the nuanced judgment necessary for mathematical content evaluation. The final quality filtering required human annotators to verify all responses and assign them to one of three categories. Responses were labeled as \textit{Accurate} when both the final answer and the reasoning steps matched the human solution. Responses received a \textit{Minor Inaccurate} label when the final answer was correct, but the reasoning contained small calculation errors or logical missteps. Finally, responses were marked as \textit{Major Inaccurate} when they contained completely incorrect reasoning or produced an incorrect final answer. Out of 998 responses, 823 were Major Inaccurate, 137 were Accurate, and 38 were Minor Inaccurate. These annotations form the foundation for assessing the LLM's mathematical reasoning and hallucination detection capabilities.}

\subsection{Evaluation Metrics}

We evaluate hallucination detection using two complementary metrics designed to capture different aspects of model performance under class imbalance conditions.

\paragraph{AUC-PR}  
We compute the Area Under the Precision–Recall Curve (AUC-PR) separately for NonFactual and Factual detection. AUC-PR is more informative than ROC-AUC under severe class imbalance, as it focuses on the performance of the minority class rather than being dominated by the majority class. Given the substantial imbalance in our AIME dataset (6:1 ratio of inaccurate to accurate responses), we interpret AUC-PR scores through relative comparisons across methods rather than absolute thresholds. High NonFactual AUC-PR indicates reliable hallucination detection, while high Factual AUC-PR reflects the ability to recognize correct outputs.

\paragraph{Ranking}  
We report the Pearson correlation coefficient (PCC) between model confidence scores and gold standard factuality labels. This metric evaluates how well models rank outputs by factuality, independent of specific classification thresholds. A strong positive correlation indicates that models consistently assign higher confidence to factual content and lower confidence to hallucinated content, which is crucial for deployment scenarios where relative confidence matters more than binary classification.

Together, these metrics provide complementary perspectives: AUC-PR captures threshold-dependent classification performance for each class, while PCC evaluates threshold-independent ranking quality. This dual evaluation ensures a comprehensive assessment of both detection sensitivity and confidence calibration across our imbalanced datasets.

\section{Method}\label{method}

Existing hallucination detection methods face a fundamental trade-off: approaches optimized for catching obvious errors often fail to validate subtle correctness, while methods excelling at recognizing accurate content frequently miss sophisticated hallucinations. To address this challenge, we introduce \textcolor{black}{SelfCheck-Eval}, a framework that evaluates three independent detection strategies, each targeting different aspects of this problem.
\textcolor{black}{An overview of the full SelfCheck-Eval framework and the three detection methods is provided in Figure \ref{fig:pipeline}.}
\textcolor{black}{The sampled responses was generated by querying the target LLM multiple times with the same query. One response serves as the target to evaluate; the others serve as sampled references. 
Each method produces a hallucination score in [0, 1].}

The Semantic \textcolor{black}{module} detects hallucinations by identifying content that deviates from expected distributional patterns, combining frequency-based analysis with semantic similarity. The Specialized Detection \textcolor{black}{module} frames the problem as natural language inference, leveraging fine-tuned LLMs to assess logical consistency between responses and evidence. The Contextual Consistency \textcolor{black}{module} exploits the reasoning capabilities of LLMs to evaluate whether generated content aligns with sampled contextual information.

Each \textcolor{black}{module} operates independently, providing complementary perspectives on the same detection task. This allows us to evaluate which approaches are most effective across different domains and types of hallucinations, as illustrated in our experimental pipeline (Figure~\ref{fig:pipeline}). The following subsections detail the technical implementation of each \textcolor{black}{module}.

\paragraph{\textbf{Semantic \textcolor{black}{module}}}

{\color{black}
The Semantic \textcolor{black}{module} operates on the assumption that hallucinated content exhibits unusual distributional patterns compared to typical responses for similar queries. While conceptually simpler than contextual embedding approaches, this frequency-based model offers computational efficiency and explicit interpretability.

Formally, given sampled generations $\{S_1,\dots,S_N\}$ and the target response $R$, we extract all tokens and build a vocabulary $V$. The target response $R$ is included in the sampled data as a smoothing mechanism, incrementing each of its token counts by one so that $R$ is contextualized within the broader distribution. For each token $t \in V$, let $\mathrm{count}(t)$ denote its frequency in the combined samples, and let
\[
\mathrm{token\_count} = \sum_{t\in V} \mathrm{count}(t)
\]
be the total number of observed tokens. We estimate smoothed unigram probabilities using Laplace smoothing with parameter $k>0$, and set $k=1$ in all experiments. The resulting smoothed probability is:
\begin{equation}
\hat{p}(t) = \frac{\mathrm{count}(t) + k}{\mathrm{token\_count} + k\,|V|}.
\end{equation}

To capture semantic variation, we extend this model using Word2Vec embeddings \cite{DBLP:journals/corr/abs-1301-3781}. For tokens $t$ and $t'$ with embedding vectors $\mathbf{v}_t$ and $\mathbf{v}_{t'}$, we compute their cosine similarity and define the semantic similarity
\[
\mathrm{similar}(t) = \{ t' \in V : \mathrm{similarity}(t,t') \ge \theta \},
\]
where $\theta = 0.9$ ensures that only highly similar tokens contribute to the aggregated probability. The semantic probability of token $t$ is then
\begin{equation}
\tilde{p}(t) = \sum_{t' \in \mathrm{similar}(t)} \hat{p}(t').
\end{equation}

The negative log-likelihood of token $t$ is defined as
\[
\mathrm{NLL}(t) = -\log \tilde{p}(t),
\]
with higher values indicating tokens that are improbable under the sampled distribution and therefore potential hallucinations.

To quantify hallucination risk at the sentence level, consider a sentence with tokens $\{t_1, \dots, t_n\}$. We compute:
\begin{equation}
\text{AvgNLL} = -\frac{1}{n}\sum_{i=1}^{n} \log\!\big(\tilde{p}(t_i)\big),
\qquad
\text{MaxNLL} = - \min_{i} \log\!\big(\tilde{p}(t_i)\big),
\end{equation}
where $n$ is the number of tokens in the sentence, AvgNLL captures overall plausibility by averaging across all tokens, while MaxNLL isolates the least probable token, highlighting localized irregularities.

At the document level, sentence-level scores are aggregated as
\begin{equation}
\mathrm{DocAvgNLL} = \frac{1}{m}\sum_{j=1}^{m} \mathrm{AvgNLL}^{(j)},
\qquad
\mathrm{DocMaxAvgNLL} = \frac{1}{m} \sum_{j=1}^{m} \mathrm{MaxNLL}^{(j)},
\end{equation}
where $m$ is the number of sentences. While the two metrics provide complementary perspectives, our evaluation adopts MaxNLL as the primary criterion, owing to its superior effectiveness in distinguishing hallucinated from factual content.

}

\paragraph{\textbf{Specialized Detection \textcolor{black}{module}}}

The Specialized Detection \textcolor{black}{module} frames hallucination detection as a natural language inference problem: if a generated sentence contradicts or is not entailed by relevant context, it likely contains hallucinations. This formulation enables transfer learning from existing NLI datasets and leverages the logical reasoning capabilities developed in these tasks.

We fine-tune LLMs on the MultiNLI~\cite{DBLP:conf/naacl/WilliamsNB18} dataset using QLoRA~\cite{DBLP:conf/nips/DettmersPHZ23} for efficient adaptation. The fine-tuning objective minimizes cross-entropy loss for the three-way classification task: entailment, neutral, and contradiction.

For hallucination detection, we assess each generated sentence against multiple context passages. \textcolor{black}{The context passages $P_j$ correspond to alternative LLM generations for the same query.} The fine-tuned LLM predicts the NLI relationship between the sentence and each context passage. We then map these predictions to factuality categories: entailment indicates accurate content, neutral suggests minor inaccuracies, and contradiction signals major inaccuracies. The final factuality score for sentence $s_i$ is computed as:
\[
S_i = \frac{1}{|P|} \sum_{j=1}^{|P|} f_{\text{score}}(s_i, P_j),
\]
where $f_{\text{score}}$ maps NLI predictions to numerical values and $|P|$ is the number of context passages.

This approach tests whether logical consistency learned from general NLI tasks transfers effectively to domain-specific hallucination detection, particularly in mathematical reasoning, where logical coherence is paramount.
\paragraph{\textbf{Contextual Consistency \textcolor{black}{module}}}

The Contextual Consistency \textcolor{black}{module} leverages the reasoning capabilities of LLMs to directly assess whether generated content is supported by available context. Unlike the previous \textcolor{black}{methods} that rely on statistical patterns or learned NLI mappings, this approach exploits the natural language understanding and reasoning abilities of LLMs themselves.
\textcolor{black}{Following Potsawee et al.\cite{DBLP:conf/emnlp/ManakulLG23}, each sentence $s_i$ in the primary LLM output is evaluated against the set of alternative passages $\{P_1,\dots,P_{|P|}\}$ associated with the same query. As outlined in Section~\ref{sec:datasets}, these passages correspond to stochastically generated responses from the same LLM prompted with the same query.}
Given a sentence $s_i$ and a sampled passage $P_j$, the LLM is queried with the following template:
\begin{quote}
\textbf{Context:} $P_j$ \\
\textbf{Sentence:} $s_i$ \\
\textbf{Question:} Is the sentence supported by the context above? \\Answer Yes or No.
\end{quote}

We consider two prompting strategies. In the \textit{zero-shot} setting, the LLM receives only the structured input above. In the \textit{chain-of-thought} setting, the prompt is extended with instructions encouraging the LLM to provide intermediate reasoning before giving a final binary answer. LLM outputs are post-processed into numerical scores: \textit{Yes} $\mapsto 0.0$ (Accurate), \textit{No} $\mapsto 1.0$ (Inaccurate), and non-conforming responses $\mapsto 0.5$ (Ambiguous). The final factuality score for sentence $s_i$ is computed as:
\[
S_i = \frac{1}{|P|} \sum_{j=1}^{|P|} f_{\text{score}}(s_i, P_j),
\]
where $f_{\text{score}}$ maps the LLM's judgment into $\{0.0, 0.5, 1.0\}$ and $|P|$ denotes the number of passages.
\textcolor{black}{For AIME dataset, we evaluate complete solutions rather than sentence-level due to the difficulty of segmenting mathematical text.}

\begin{table}[htbp]
\centering
\small
\renewcommand{\arraystretch}{1.1}

\resizebox{\textwidth}{!}{%
\begin{tabular}{>{\raggedright\arraybackslash}m{3.8cm} p{2.2cm} p{1.0cm} p{1.0cm} ccc ccc}

\toprule
\textbf{Method} & \textbf{LLM} & \textbf{Size} & \textbf{Prompt} &
\multicolumn{3}{c}{\textbf{WikiBio Dataset}} &
\multicolumn{3}{c}{\textbf{AIME Dataset}} \\
\cmidrule(lr){5-7} \cmidrule(lr){8-10}
& & & & \textbf{NonFact} & \textbf{Factual} & \textbf{Ranking} &
\textbf{NonFact} & \textbf{Factual} & \textbf{Ranking} \\
\midrule
\makecell[l]{SelfCheckGPT \\ (Unigram)}        & -            & -    & -   & 85.63 & 58.47 & 64.71 & 85.69 & 14.91 & -    \\
\makecell[l]{SelfCheckGPT \\ (BertScore)}      & -            & -    & -   & 85.63 & 58.47 & 64.71 & \textbf{88.51} & \textbf{17.82} & - \\
\makecell[l]{Semantic \\ \textcolor{black}{module}}                & -            & -    & -   & \textbf{86.97} & \textbf{59.02} & \textbf{65.88} & 87.24 & 14.62 & 11.94 \\
\midrule[2.2pt]
\makecell[l]{SelfCheckGPT \\ (Fine-tuned)}     & DeBERTa-v3   & -    & -   & 92.50 & 66.08 & 74.14 & 87.64 & 17.45 & 5.32 \\
\midrule
\multirow{10}{3.8cm}{\makecell[l]{Specialized \\ Detection \textcolor{black}{module} \\ (pre-trained)}}
 & Roberta-large & 340M & ZS  & 76.44 & 31.20 &  6.07 & -     & -     & -    \\
 & Phi-3         & 3.8B & ZS  & \textbf{89.27} & \textbf{56.26} & \textbf{61.88} & 90.34 & 19.16 & 12.75 \\
 & Llama 3.1     & 8B   & ZS  & 78.90 & 37.12 & 34.25 & 87.76 & 21.10 & 3.16 \\
 & Mistral       & 7B   & ZS  & 86.13 & 58.60 & 41.15 & 91.33 & 17.11 & 2.57 \\
 & Gemma         & 7B   & ZS  & 75.41 & 30.87 &  4.80 & \textbf{93.38} & \textbf{20.38} & \textbf{21.76} \\
 & Phi-3         & 3.8B & CoT & 75.08 & 29.67 & 13.79 & \underline{87.76} & \underline{21.10} & \underline{3.16} \\
 & Llama 3.1     & 8B   & CoT & 63.88 & 21.75 & -     & 85.33 & 13.13 & -    \\
 & Mistral       & 7B   & CoT & \underline{81.91} & \underline{46.10} & \underline{43.49} & 91.33 & 17.11 & 2.57 \\
 & Gemma         & 7B   & CoT & 70.97 & 25.99 &  -    & 90.07 & 13.00 & 0.50 \\
 & Roberta-large & 340M & CoT & 73.86 & 31.47 & 13.02 & -     & -     & -    \\
\midrule
\multirow{4}{3.8cm}{\makecell[l]{Specialized \\ Detection \textcolor{black}{module} \\ (fine-tuned)}}
 & Phi-3       & 3.8B & -   & \underline{92.87} & \underline{65.25} & \underline{73.54} & \textbf{93.38} & \textbf{20.38} & \textbf{21.76} \\
 & Llama 3.1   & 8B   & -   & 76.85 & 29.71 &  8.91 & 79.63 &  9.74 & -    \\
 & Mistral     & 7B   & -   & \textbf{92.68} & \textbf{67.10} & \textbf{75.63} & \underline{92.91} & \underline{17.37} & \underline{20.05} \\
 & Gemma       & 7B   & -   & 83.47 & 43.12 & 50.98 & 82.58 & 10.70 & -    \\
\midrule[2.2pt]
\multirow{2}{3.8cm}{\makecell[l]{SelfCheckGPT \\ (Prompt-based)}}
 & Mistral     & 7B   & ZS  & 91.31 & 62.76 & 74.46 & 92.15 & 45.31 & 17.19 \\
 & gpt3.5      & -    & ZS  & 93.42 & 67.09 & 78.32 & \textbf{95.66} & \underline{56.95} & \underline{30.08} \\
\midrule
\multirow{6}{3.8cm}{\makecell[l]{Contextual \\ Consistency \textcolor{black}{module} \\ (Prompt-based)}}
 & Llama 3.1   & 8B   & ZS  & 92.85 & 70.73 & 76.54 & 94.79 & 37.75 & 29.67 \\
 & gpt4o       & -    & ZS  & \textbf{94.00} & \textbf{74.11} & \underline{77.48} & 93.93 & 24.26 & 21.13 \\
 & Mistral     & 7B   & CoT & 91.74 & 64.01 & 75.40 & 93.63 & 37.98 & 22.51 \\
 & Llama 3.1   & 8B   & CoT & \underline{93.64} & \underline{70.26} & \textbf{78.48} & 91.77 & 24.89 & 14.23 \\
 & gpt3.5      & -    & CoT & 90.59 & 62.11 & 72.17 & \underline{95.28} & \textbf{57.62} & 25.42 \\
 & gpt4o       & -    & CoT & 94.14 & 74.95 & 76.33 & 94.89 & 30.58 & \textbf{30.68} \\
\bottomrule
\end{tabular}%
}
\caption{Consolidated comparison of methods, LLMs sizes, and prompting strategies on the WikiBio and AIME datasets. Reported metrics include NonFactual (AUC-PR), Factual (AUC-PR), and Ranking (PCC). The best result is shown in bold, and the second-best is underlined.}
\label{tab:consolidated_comparison}
\end{table}

\section{Experiments}
\subsection{Setup}
\paragraph{LLMs} 
We evaluate our approach using a diverse set of LLMs, covering both proprietary and open-source systems. The selected LLMs include GPT-4~\cite{DBLP:journals/corr/abs-2303-08774}, GPT-3.5~\cite{DBLP:journals/corr/abs-2303-10420}, Phi-3-mini~\cite{abdin2024phi}, Llama 3.1 (8B)~\cite{dubey2024llama}, Mistral (7B)~\cite{jiang2023mistral}, and Gemma (7B)~\cite{team2024gemma}. For benchmarking, we also include RoBERTa-large~\cite{DBLP:journals/corr/abs-1907-11692}, which provides a strong reference point on natural language understanding tasks. This selection enables evaluation of hallucination detection across a broad spectrum of LLM families and capacities.  

\paragraph{Implementation Details} 
For GPT-4 and GPT-3.5, we rely on OpenAI’s API (December 2024). Responses are generated with a temperature of $0.6$, balancing diversity and consistency, and a maximum token limit of $2048$ to ensure efficiency. All experiments are conducted on a computational setup comprising one NVIDIA A500 GPU and two NVIDIA 2080Ti GPUs. For comparability, temperature and decoding settings are standardized across LLMs.  

\paragraph{NLI Fine-tuning.} 
We fine-tuned LLaMA~3.1~8B~\cite{dubey2024llama}, Mistral~7B~\cite{jiang2023mistral}, Gemma~7B~\cite{team2024gemma}, and Phi-3-mini~\cite{abdin2024phi} on the MultiNLI dataset~\cite{DBLP:conf/naacl/WilliamsNB18}. All LLMs were \textcolor{black}{fine-tuned} with QLoRA 4-bit quantization and LoRA adapters ($r=16$, $\alpha=32$, dropout $0.05$). \textcolor{black}{Fine-tuning} was performed for three epochs with a learning rate of $2\cdot10^{-4}$, batch size $8$, and AdamW as optimizer.

\subsection{Evaluation Methods}

\paragraph{Semantic \textcolor{black}{module}} 
The \textit{Semantic \textcolor{black}{module}} estimates lexical similarity between generated responses and sampled passages using semantic $n$-gram overlap. It provides a lightweight, frequency-based baseline for hallucination detection.  

\paragraph{Specialized Detection \textcolor{black}{module}} 
The \textit{Specialized Detection \textcolor{black}{module}} is implemented in two variants:  
(i) an NLI-based prompting approach, where pre-trained LLMs are queried without fine-tuning; and  
(ii) fine-tuned LLMs on the MultiNLI dataset to improve sentence-level factuality judgments. Both variants enable entailment–contradiction reasoning for hallucination detection.  

\paragraph{Contextual Consistency \textcolor{black}{module}} 
The \textit{Contextual Consistency \textcolor{black}{module}} evaluates whether a sentence is supported by retrieved context passages. We test two prompting strategies: zero-shot (ZS), where the LLMs directly answer based on structured input, and chain-of-thought (CoT), which encourages intermediate reasoning before the final answer.  

\paragraph{SelfCheckGPT Baseline} 
For comparison, we include SelfCheckGPT~\cite{DBLP:conf/emnlp/ManakulLG23}, a black-box hallucination detection method that samples multiple generations from an LLM and evaluates factual alignment across them. This established approach serves as our primary comparison baseline for validating the effectiveness of our proposed \textcolor{black}{modules}. Specifically, we compare against Unigram, BERTScore, SelfCheckGPT-Fine-tuned, and SelfCheckGPT-Prompt.

\subsection{Main Results}
\label{main_results}

We evaluate the effectiveness of \textit{\textcolor{black}{SelfCheck-Eval}} on the WikiBio dataset and on AIME-Math, a novel benchmark proposed in this work. 
Table~\ref{tab:consolidated_comparison} reports the overall performance across the three \textcolor{black}{modules} introduced in Section~\ref{method}: the Semantic \textcolor{black}{module}, the Specialized Detection \textcolor{black}{module}, and the Contextual Consistency \textcolor{black}{module}. 
The following subsections provide a detailed analysis of their individual contributions.

\paragraph{Semantic \textcolor{black}{module}}

The Semantic \textcolor{black}{module}, which combines frequency-based unigram probabilities with Word2Vec semantic similarity, demonstrates domain-dependent effectiveness with markedly different performance characteristics across the two benchmarks.

On the WikiBio dataset, the Semantic \textcolor{black}{module} achieves consistent but incremental improvements over frequency-based baselines. For NonFactual detection, it reaches an AUC-PR of 86.97, a 1.34-point improvement over the SelfCheckGPT Unigram baseline (85.63). The gains are more modest on Factual detection, with 59.02 compared to 58.47 for both Unigram and BertScore variants, a difference of just 0.55 points. Ranking correlation shows a similar pattern, reaching 65.88 versus 64.71 for the baselines, representing a 1.17-point increase. The results suggest that Word2Vec semantic enhancement provides some benefit over pure frequency-based approaches in biographical text, though the gains are limited compared to the computational overhead of the semantic similarity calculations.

The results change dramatically when transitioning to the AIME dataset. While NonFactual detection remains relatively stable at 87.24 (compared to 85.69 for the \textcolor{black}{u}nigram baseline), both Factual detection and Ranking performance collapse severely. Factual detection drops to just 14.62, compared to 59.02 on WikiBio, representing one of the most severe performance drops observed across domains, while ranking correlation falls to 11.94. In comparison, SelfCheckGPT with BertScore achieves superior performance on AIME with 88.51 NonFactual and 17.82 Factual detection, suggesting that BERT's contextual embeddings capture aspects of mathematical reasoning that Word2Vec's static representations miss.

This substantial performance degradation highlights a fundamental limitation of frequency-based approaches in mathematical domains. While the Semantic \textcolor{black}{module} can effectively identify distributional anomalies in narrative text where unusual word choices often signal factual errors, it struggles with mathematical content where correctness depends on logical relationships rather than lexical patterns. The method's reliance on surface-level similarity makes it poorly suited for detecting the subtle logical inconsistencies that characterize mathematical hallucinations.

\paragraph{Specialized Detection \textcolor{black}{module}} 

The Specialized Detection \textcolor{black}{module} demonstrates a consistent pattern across both pre-trained and fine-tuned variants: strong contradiction detection but weak correctness validation, with this asymmetry becoming extreme in mathematical domains.

Pre-trained LLMs show variable performance depending on architecture and prompting strategy. On WikiBio, zero-shot prompting generally outperforms chain-of-thought, with Phi-3 3.8B achieving the best balance (89.27 NonFactual, 56.26 Factual). LLMs like Mistral 7B maintain competitive Factual detection (58.60) but struggle with ranking tasks. The transition to AIME reveals the method's core limitation: while NonFactual detection remains robust across LLMs (85-93\%), Factual detection collapses uniformly, with even the best performer (Gemma 7B) reaching only 20.38\%.

Fine-tuning on MultiNLI substantially improves WikiBio performance, with Mistral 7B and Phi-3 3.8B both exceeding 92\% NonFactual and 65\% Factual detection. However, this supervised approach fails to transfer to mathematical reasoning. On AIME, fine-tuned LLMs replicate the same asymmetric pattern: excellent contradiction detection ($>$90\%) paired with poor correctness validation ($<$22\%).

This consistent asymmetry across \textcolor{black}{fine-tuning} paradigms suggests that current NLI approaches, when \textcolor{black}{fine-tuned}  on general domain datasets like MultiNLI, are fundamentally limited in their ability to validate mathematical reasoning, regardless of LLMs scale or \textcolor{black}{fine-tuning strategy}.
\paragraph{Contextual Consistency \textcolor{black}{module}}

The Contextual Consistency \textcolor{black}{module} achieves the strongest performance among all methods tested, establishing new benchmarks on WikiBio while simultaneously revealing the persistent challenges of mathematical reasoning validation.

On WikiBio, this \textcolor{black}{module} delivers state-of-the-art results across all metrics. GPT-4o sets the highest Factual detection at 74.95\% (CoT) and 74.11\% (ZS), substantially outperforming all previous methods, including fine-tuned baselines. NonFactual detection consistently exceeds 91\% across LLMs, with GPT-4o CoT reaching 94.14\%. Even open-weight LLMs achieve competitive performance, with LLaMA 3.1 reaching 70.73\% Factual detection\textcolor{black}{, outperforming SelfCheckGPT's best configuration with GPT-3.5 (67.09\%)}. Ranking correlations similarly excel, with LLaMA 8B with CoT achieving the highest score of 78.48\%.

Despite this clear superiority in biographical text, the mathematical domain exposes fundamental limitations that persist even for the best-performing approach. While NonFactual detection remains robust (91-96\%), Factual detection undergoes severe degradation. Most notably, GPT-4o, the clear leader on WikiBio, drops dramatically to 24.26\% (ZS) and 30.58\% (CoT).\textcolor{black}{In contrast, our framework with GPT-3.5 achieves 57.62\% Factual 
detection (CoT), closely aligning with the SelfCheckGPT zero-shot baseline  (56.95\%), and substantially higher than other prompt-based variants.}  Open-weight LLMs cluster between 24-38\%, maintaining the consistent pattern observed across all \textcolor{black}{methods}.

These results establish the Contextual Consistency \textcolor{black}{module} as the most effective approach for general hallucination detection, while confirming that even the best-performing methods face severe limitations when applied to mathematical reasoning tasks. The persistent domain gap suggests that the challenge lies not in the sophistication of the detection method, but in fundamental aspects of mathematical validation that current approaches cannot address.

\paragraph{Overall Comparison}
{\color{black}

Table~\ref{tab:consolidated_comparison} reveals several cross-method trends that 
highlight the underlying challenges of hallucination detection across domains.
First, all methods show a pronounced asymmetry between NonFactual and Factual 
detection on both datasets. NonFactual detection remains consistently high 
(85--96\%) regardless of LLM scale or architecture, suggesting that identifying 
explicit contradictions or distributional anomalies is comparatively easy. In 
contrast, Factual detection is substantially lower and strongly domain-dependent, 
indicating that validating correctness requires deeper semantic or logical 
reasoning that current approaches struggle to capture.

Second, the domain gap between WikiBio and AIME is systematic and affects every 
category of method. Whether frequency-based, NLI-based, or prompt-based, all 
methods exhibit a sharp degradation in Factual detection on AIME, often a 
40--55 point drop. This suggests that the difficulty is intrinsic to 
mathematical reasoning, where correctness hinges on domain-specific logical 
relations rather than surface-level lexical cues.

Third, the ranking metric mirrors the same pattern. LLMs that perform well on 
WikiBio (such as GPT-4o or LLaMA 3.1 in the prompt-based setting) experiences dramatic drops in ranking correlation on AIME, indicating that even 
when contradictions are detectable, LLMs fail to organize mathematical 
solutions by correctness. This further confirms that existing semantic 
representations are insufficient for structured reasoning tasks.

Finally, the relative performance gap between methods narrows considerably on AIME. The advantages of contextual or semantic embeddings largely disappear, and even the best-performing LLMs converge to a similarly low ceiling of 20--38\% Factual detection, underscoring the difficulty of detecting hallucinations in mathematical reasoning.
}

\begin{figure}[h]
    \centering
    \includegraphics[width=0.34\textwidth]{bar-graph.png} 
    \caption{Analysis of LLM capacity impact.}
    \label{fig:LLM_capa_ansl}
\end{figure}

\subsection{Experimental Analysis}

\paragraph{Examining the Influences of \textcolor{black}{module} Capacity}

Figure~\ref{fig:LLM_capa_ansl} shows that increasing LLM capacities, especially in specialized detection \textcolor{black}{module}, significantly improves hallucination detection. Specialized detection \textcolor{black}{module} such as Phi-3 and Mistral outperform their pre-trained counterparts. Table~\ref{tab:consolidated_comparison} highlights performance variations across \textcolor{black}{module} capacities, with specialized detection \textcolor{black}{module} consistently surpassing other methods, achieving top NonFactual, Factual, and Ranking scores on both WikiBio and AIME datasets. Among contextual consistency \textcolor{black}{module}, GPT-4o with CoT prompting and GPT-3.5 in zero-shot prompting show strong performance, emphasizing the importance of LLM size and fine-tuning in hallucination detection.

\paragraph{Challenges and Anomalies in Performance}

The specialized detection \textcolor{black}{module} analysis reveals some interesting findings. On the AIME dataset, Roberta-large fails in pre-trained settings due to its limited context window. Also in the AIME dataset, significant gaps between NonFactual (AUC-PR), Factual (AUC-PR), and Ranking (PCC) scores raised the dataset’s imbalance, caused because of the LLMs' limitation in the complex mathematics domain. Notably, Gemma (7B) in ZS pre-trained LLM matches Phi-3 (3.8B) fine-tuned scores, likely due to Gemma’s robust architecture and the dataset’s imbalanced nature, which limits sensitivity to LLM-specific optimizations.

\subsection{Analysis for Mathematical domain}
\label{analysis-math}

The dramatic performance drop observed when transitioning from WikiBio to AIME across all \textcolor{black}{SelfCheck-Eval} components (Section~\ref{main_results}) raises a fundamental question: are these failures inherent to black-box approaches, or can specialized \textcolor{black}{fine-tuning} paradigms overcome the mathematical reasoning challenge?
We evaluate three \textcolor{black}{fine-tuning} strategies targeting different aspects of mathematical reasoning: logical inference (LogiQA 2.0 \cite{logiqa}), preference learning (RewardMath \cite{rewardmath}), and process supervision (PRM800K \cite{lightman2023letsverify}, MathShepherd \cite{wang2024mathshepherd}). 
Following \textcolor{black}{ Li et al.}~\cite{fgprm2024}, we frame hallucination detection as a binary classification task and report precision, recall, and F1-score as primary metrics. While \textcolor{black}{ Li et al.}~\cite{fgprm2024} focuses on a taxonomy of hallucination types, our goal is to examine how different paradigms transfer to mathematical reasoning for the fundamental binary distinction between hallucinated and non-hallucinated outputs. Our analysis disentangles performance across the two classes (hallucinated vs non-hallucinated), providing clearer insights into LLM behavior. For consistency with prior evaluations, we also report AUC scores.
\subsubsection{NLI-based Fine-tuning of LLMs for Hallucination Detection (LogiQA 2.0)}

\begin{table*}[h]
\centering
\resizebox{\textwidth}{!}{%
\begin{tabular}{lcccccccc}
\hline
\textbf{LLM} & \textbf{P(Hallu)} & \textbf{R(Hallu)} & \textbf{F1(Hallu)} & 
\textbf{P(Non-Hallu)} & \textbf{R(Non-Hallu)} & \textbf{F1(Non-Hallu)} & 
\textbf{AUC(Hallu)} & \textbf{AUC(Non-Hallu)} \\
\hline
Llama3-8B        & \textbf{0.94} & 0.52 & 0.67 & \textbf{0.21} & 0.80 & \textbf{0.33} & \textbf{93.26} & 22.60 \\
Qwen2.5-7B       & 0.81 & 0.59 & 0.68 & 0.06 & 0.15 & 0.08 & 79.45 & 10.22 \\
Qwen2.5-Math-7B  & 0.86 & \textbf{0.89} & \textbf{0.87} & 0.12 & 0.09 & 0.11 & 84.19 & 12.29 \\
Phi-3            & 0.91 & 0.23 & 0.36 & 0.15 & \textbf{0.86} & 0.26 & 90.42 & \textbf{24.01} \\
\hline
\end{tabular}}
\caption{Evaluation results of LLMs on hallucination detection for the logic NLI task. Metrics include precision (P), recall (R), and F1 for both hallucination and non-hallucination classes, as well as AUC scores (in \%).}

\label{tab:nli-logicqa}
\end{table*}

Mathematical errors are fundamentally logical inconsistencies that can be detected through entailment reasoning.
Logical inference represents a natural candidate for transfer learning in hallucination detection: both require distinguishing valid from invalid reasoning. We fine-tune four LLMs on LogiQA 2.0~\cite{logiqa}, a large-scale NLI dataset designed to capture logical patterns beyond single sentences, and test their transfer to mathematical hallucination detection. A detailed description of the datasets and the fine-tuning parameters can be found in Appendix~\ref{appendix:logicqa}

The results (Table~\ref{tab:nli-logicqa}) reveal both the potential and the limitations of NLI-based transfer. All LLMs achieve high AUC values for hallucination classification (79–93\%), with F1-scores ranging from 0.36 (Phi-3) to 0.87 (Qwen2.5-Math). However, performance deteriorates sharply for the non-hallucination class: AUC values fall below 25\% and F1-scores remain under 0.35. This indicates that NLI \textcolor{black}{fine-tuning} promotes a conservative detection strategy in which LLMs default to labeling uncertain cases as hallucinated.
The behavior varies across architectures. LLaMA3-8B provides the most balanced transfer, with very high precision for hallucination detection (0.94) and the strongest non-hallucination F1-score (0.33). Qwen2.5-Math-7B is highly sensitive to hallucinations (F1 = 0.87, recall = 0.89) but almost unable to validate correct reasoning (F1 = 0.11). Phi-3 takes the opposite stance, showing high non-hallucination recall (0.86) but very low hallucination recall (0.23).
While NLI-based \textcolor{black}{fine-tuning} equips LLMs with strong contradiction detection, it does not transfer to validating correctness in mathematical reasoning.
\subsubsection{Reward Models for Hallucination Detection}

\begin{table*}[h]
\centering
\renewcommand{\arraystretch}{1.15}
\resizebox{\textwidth}{!}{%
\begin{tabular}{l c c c c c c c c}
\hline
\textbf{LLM} &
\makecell{\textbf{P}\\\textbf{(Hallu)}} &
\makecell{\textbf{R}\\\textbf{(Hallu)}} &
\makecell{\textbf{F1}\\\textbf{(Hallu)}} &
\makecell{\textbf{P}\\\textbf{(Non-Hallu)}} &
\makecell{\textbf{R}\\\textbf{(Non-Hallu)}} &
\makecell{\textbf{F1}\\\textbf{(Non-Hallu)}} &
\makecell{\textbf{AUC}\\\textbf{(Hallu)}} &
\makecell{\textbf{AUC}\\\textbf{(Non-Hallu)}} \\
\hline
Mistral-7B        & 0.85 & 0.36 & 0.50 & 0.13 & 0.61 & 0.22 & 86.50 & 13.39 \\
Qwen2.5-7B        & 0.91 & 0.19 & 0.32 & 0.15 & 0.88 & 0.25 & \textbf{88.30} & 13.92 \\
Qwen2.5-Math-7B   & 0.86 & 0.70 & 0.77 & 0.14 & 0.31 & 0.20 & 86.47 & 15.67 \\
Llama3-8B         & 0.87 & 0.89 & 0.88 & 0.21 & 0.18 & 0.19 & 88.07 & \textbf{16.58} \\
\hline
\end{tabular}%
}
\caption{Evaluation results of fine-tuned LLMs on RewardMATH on hallucination detection. Metrics include precision (P), recall (R), and F1 for both hallucination and non-hallucination classes, as well as AUC (\%).}
\label{tab:results_reward_model}
\end{table*}

Preference-based fine-tuning can capture the subtle quality differences between valid and invalid mathematical reasoning.
Reward modeling has emerged as a key paradigm for aligning LLMs with human expectations, as it leverages preference signals rather than categorical labels~\cite{rewardmodel-openai}. This makes it particularly relevant for hallucination detection, where errors are not always strictly factual but often involve issues of plausibility, coherence, or logical soundness.
The RewardMath framework~\cite{rewardmath} extends this principle to mathematical reasoning, where hallucinations typically take the form of invalid steps or unsupported conclusions. In this setting, reward models act as evaluators of reasoning quality, distinguishing between valid and invalid solution paths and enabling a finer-grained form of hallucination detection that goes beyond surface correctness.

A description of the dataset structure, fine-tuning parameters, and illustrative examples can be found in Appendix~\ref{appendix:rewardmath}.

The results (Table~\ref{tab:results_reward_model}) show that the reward-based approach achieves consistently high AUC values for hallucination detection (86.5–88.3\%). LLaMA3-8B obtains the strongest hallucination classification (F1 = 0.88), followed by Qwen2.5-Math-7B (0.77). In contrast, Mistral-7B (0.50) and Qwen2.5-7B (0.32) perform considerably worse. Performance on the non-hallucination class, however, remains weak: precision ranges from 0.13 to 0.21, recall varies widely (0.18–0.88), and F1 never exceeds 0.25. AUC values for the non-hallucination class also remain low (13.4–16.6\%).
Reward-based LLMs are effective at flagging flawed reasoning but struggle to reliably recognize correct solutions, reproducing the same asymmetric pattern observed with NLI-based approaches.
\subsubsection{PRM for Hallucination Detection}

Step-level supervision can address error propagation in multi-step mathematical reasoning.
Reward models assess the overall quality of generated outputs, but they remain limited when reasoning unfolds across multiple steps. In mathematical problem solving, errors often emerge gradually in intermediate steps rather than appearing only in the final answer. Process Reward Models (PRMs) address this limitation by assigning evaluative signals at the level of intermediate reasoning steps, thereby complementing or replacing outcome-based supervision.
The idea was first introduced by Uesato et al.\cite{uesato2022process}, showing that rewarding intermediate correctness improves reasoning reliability. Subsequent work, most notably \cite{lightman2023letsverify}, formalized this principle by providing a large-scale dataset of human-annotated step-level feedback, namely PRM800K, and demonstrated that process supervision mitigates compounding errors in multi-step reasoning.
This approach is especially relevant to our setting, where AIME solutions are naturally decomposed into step-by-step traces. Hallucinations often take the form of subtle errors in intermediate transitions rather than overt mistakes in the final result. PRMs make it possible to estimate the probability that each reasoning step introduces a hallucination and to aggregate these signals into a structured diagnostic of reasoning quality.

To evaluate PRMs for hallucination detection, we fine-tuned LLMs on two complementary datasets: PRM800K~\cite{lightman2023letsverify}, which provides step-level human annotations of candidate solutions, and Math-Shepherd~\cite{wang2024mathshepherd}, which automatically generates process-level labels for large-scale supervision. Both datasets are designed to capture intermediate reasoning quality rather than only final answers, making them well-suited to process-level hallucination detection. Further details on dataset statistics, fine-tuning settings, and example instances are reported in Appendix~\ref{appendix:prm800k} and ~\ref{appendix:mathshepherd}. The results in Table~\ref{tab:results_reward_model_prm} show that PRM fine-tuning yields strong performance on hallucination detection, with F1(Hallu) approaching or exceeding 0.90 in several settings. LLaMA3-8B on PRM800K reaches 0.88, Qwen2.5-Math-7B on PRM800K attains 0.93, and LLaMA3-8B on Math-Shepherd achieves 0.93. High recall for the hallucination class, often close to 1.0 as in Qwen2.5-Math-7B on PRM800K and LLaMA3-8B on Math-Shepherd, confirms that PRMs are highly effective at flagging flawed reasoning steps.

However, this strength comes at the expense of non-hallucination detection. Across all F1(Non-Hallu) remains very low, never exceeding 0.22, with the best result obtained by Qwen2.5-7B on Math-Shepherd. Precision for the non-hallucination class is nearly zero in several configurations, for instance, LLaMA3-8B on PRM800K with 0.01. This imbalance suggests that PRMs adopt an overly conservative strategy, systematically favoring hallucination classification and producing frequent false positives when evaluating valid reasoning.

Process-level supervision, therefore, enables near-perfect sensitivity to hallucinated reasoning but struggles to develop complementary capabilities for validating correct solutions, representing the most extreme manifestation of the conservative bias observed across all approaches.

\begin{table*}[h]
\centering
\renewcommand{\arraystretch}{1.15}
\resizebox{\textwidth}{!}{%
\begin{tabular}{l c c c c c c c c}
\hline
\textbf{LLM} &
\makecell{\textbf{P}\\\textbf{(Hallu)}} &
\makecell{\textbf{R}\\\textbf{(Hallu)}} &
\makecell{\textbf{F1}\\\textbf{(Hallu)}} &
\makecell{\textbf{P}\\\textbf{(Non-Hallu)}} &
\makecell{\textbf{R}\\\textbf{(Non-Hallu)}} &
\makecell{\textbf{F1}\\\textbf{(Non-Hallu)}} &
\makecell{\textbf{AUC}\\\textbf{(Hallu)}} &
\makecell{\textbf{AUC}\\\textbf{(Non-Hallu)}} \\
\hline
Llama3-8B (PRM800k)            & 0.85 & 0.92 & 0.88 & 0.01 & 0.07 & 0.01 & 79.25 & 9.44 \\
Qwen2.5-Math-7B (PRM800k)      & 0.86 & \textbf{1.00} & \textbf{0.93} & 0.00 & 0.00 & 0.00 & 81.13 & 9.94 \\
Qwen2.5-7B (PRM800k)           & 0.85 & 0.88 & 0.87 & 0.03 & 0.02 & 0.02 & \textbf{83.36} & 10.75 \\
Qwen2.5-Math-7B (MathShepherd) & 0.76 & 0.14 & 0.24 & 0.12 & \textbf{0.72} & 0.20 & 82.52 & \textbf{13.89} \\
Qwen2.5-7B (MathShepherd)      & 0.83 & 0.20 & 0.33 & 0.13 & 0.74 & \textbf{0.22} & 82.89 & 10.25 \\
Llama3-8B (MathShepherd)       & \textbf{0.86} & 1.00 & 0.93 & \textbf{0.50} & 0.01 & 0.01 & 81.36 & 10.61 \\
\hline
\end{tabular}%
}

\caption{Evaluation results of PRM fine-tuned LLMs (MathShepherd and PRM800k) for hallucination detection. Best results for each metric are highlighted in bold. Metrics include precision (P), recall (R), and F1 for both hallucination and non-hallucination classes, as well as AUC (\%).}
\label{tab:results_reward_model_prm}
\end{table*}

\subsubsection{The Mathematical Domain Challenge}

Across all three specialized paradigms, NLI fine-tuning, reward modeling, and process supervision, we observe a consistent pattern of asymmetric performance: while LLMs achieve high or even near-perfect sensitivity to hallucinated content (F1 for hallucination detection often $>0.9$), they consistently struggle with validating correct reasoning (F1 for non-hallucination detection rarely exceeds 0.25, with most results below 0.20). This mirrors the behavior observed in \textcolor{black}{SelfCheck-Eval}, suggesting that the mathematical domain poses fundamental challenges that require better training methodology or architectural choice.

When we compare these specialized approaches to our original \textcolor{black}{module} on AIME, the pattern is remarkably consistent. The Contextual Consistency \textcolor{black}{module} achieved 30.58\% factual detection with GPT-4o, while our best specialized approaches reach similar levels (such as Qwen2.5-Math-7B on RewardMath with F1 = 0.20 for non-hallucinations). This convergence across fundamentally different approaches suggests the limitation is domain-inherent rather than method-specific.

Interestingly, LLMs specialized in mathematical reasoning do not perform better on the non-hallucination class under process supervision. Qwen2.5-Math-7B, for example, achieves near-perfect recall for hallucinations (1.0 on PRM800K) but entirely fails to recognize correct reasoning steps (F1(Non-Hallu) = 0.00). By contrast, its non-specialized counterpart Qwen2.5-7B, despite weaker hallucination performance, reaches a higher balance with F1(Non-Hallu) = 0.22 on Math-Shepherd. This suggests that mathematical specialization amplifies the conservative bias of PRMs: rather than learning to validate correct reasoning, specialized LLMs default to classifying nearly all cases as hallucinations.

The failure appears rooted in the fundamental difference between fine-tuning domains and mathematical reasoning tasks. LLMs fine-tuned on logical inference datasets like LogiQA 2.0 learn to identify contradictions in natural language, but these skills do not transfer effectively to mathematical content, where correctness depends on precise logical relationships and multi-step reasoning chains. Reward models trained on mathematical solution preferences can distinguish quality differences but struggle with subtle reasoning errors. Process reward models, despite step-level supervision, adopt overly conservative strategies that flag valid reasoning as potentially problematic.

Our findings demonstrate that current approaches face significant limitations when applied to mathematical reasoning, regardless of the fine-tuning paradigm employed. The consistent performance ceiling across NLI fine-tuning, reward modeling, and process supervision suggests that effective mathematical hallucination detection may require fundamentally different methods explicitly designed for formal reasoning domains—potentially integrating symbolic verification, proof-checking, or hybrid neuro-symbolic techniques. The convergence of results across diverse fine-tuning paradigms thus represents a robust finding about the inherent difficulty of applying statistical learning methods to domains that demand rigorous logical validation.

\section{Conclusions}
We introduce \textcolor{black}{SelfCheck-Eval}, a systematic framework for evaluating hallucination detection across domains, and the AIME Math Hallucination benchmark for mathematical reasoning evaluation. Our analysis reveals fundamental domain-dependent limitations in current approaches.
Our evaluation demonstrates that mathematical reasoning poses unique challenges for hallucination detection. While all methods achieve robust performance on biographical content, they exhibit consistent asymmetric behaviour in mathematical domains. In fact, all methods are able to detect incorrect solutions while systematically failing to validate correct reasoning. This pattern persists across NLI fine-tuning, preference learning, and process supervision, affecting even mathematics-specialised LLMs.
The consistent performance convergence across diverse fine-tuning paradigms suggests that mathematical correctness requires rigorous logical validation that transcends current statistical learning approaches. Unlike biographical facts validated through distributional patterns, mathematical reasoning demands domain-specific verification mechanisms.

\section{Limitations}
Our evaluation focuses on competition-level mathematical reasoning through AIME problems versus biographical content from WikiBio. While this comparison reveals fundamental differences between domains, the generalizability to other formal reasoning contexts such as physics, computer science, and symbolic logic requires further investigation. The AIME problems represent competition-level mathematics requiring creative problem-solving strategies, which may not reflect all mathematical reasoning scenarios encountered in practical applications. Additionally, our binary classification approach (hallucinated vs. non-hallucinated) may not capture the nuanced spectrum of mathematical errors, from minor computational mistakes to fundamental conceptual misunderstandings.
\textcolor{black}{Future work could incorporate additional evaluation metrics beyond AUC-PR and PCC to enable a more comprehensive assessment of performance.}
Our findings suggest several critical research directions: developing hybrid approaches that integrate statistical detection with symbolic verification for mathematical domains, extending evaluation to broader formal reasoning contexts, and investigating domain-adaptive architectures specifically designed for rigorous logical validation. 

\section*{RESOURCE AVAILABILITY}


\subsection*{Lead contact}


Requests for further information and resources should be directed to and will be fulfilled by the lead contact, Gollam Rabby (gollam.rabby@l3s.de).




\subsection*{Data and code availability}
The dataset utilized for this study is accessible through Hugging Face. Interested readers and researchers can obtain the dataset by visiting the following links: 1. (\url{https://huggingface.co/datasets/tourist800/AIME_Hallucination_Detection}), 2. \url{https://huggingface.co/datasets/potsawee/wiki_bio_gpt3_hallucination}. The study was carried out exclusively using open-source software packages. All scripts, outcomes, post-processed datasets, and features will be accessible to the public at \url{https://github.com/DIYANAPV/\textcolor{black}{SelfCheck}}.

\section*{ACKNOWLEDGMENTS}


We acknowledge the support of the KISSKI project (funding no. 01IS22093C) for providing computational resources, which will enable us to extend this research in the future. Giusy Giulia Tuccari acknowledges FOSSR (Fostering Open Science in Social Science Research), funded by the European Union-NextGenerationEU under NRRP Grant agreement n. MUR IR0000008.

\section*{AUTHOR CONTRIBUTIONS}

This work was carried out through close collaboration among all authors. G.R. designed the experiment, led the experiment, and wrote the manuscript. D.M. and G.G.T. were responsible for implementing, conducting the experiments, and writing the manuscript. S.A. and S.V. played a significant role in conceiving the experiment design and contributed to the writing of the manuscript.






\section*{DECLARATION OF GENERATIVE AI AND AI-ASSISTED TECHNOLOGIES}


During the preparation of this work, the author(s) used ChatGPT in order to improve grammatical clarity and correct language errors. After using this tool, the author(s) reviewed and edited the content as needed and take(s) full responsibility for the content of the publication.

\bibliography{references}

@inproceedings{DBLP:conf/ijcai/WangSORRE24,
  author = {Xindi Wang and Mahsa Salmani and Parsa Omidi and Xiangyu Ren and Mehdi Rezagholizadeh and Armaghan Eshaghi},
  title = {Beyond the Limits: {A} Survey of Techniques to Extend the Context Length in Large Language Models},
  booktitle = {Proceedings of the Thirty-Third International Joint Conference on Artificial Intelligence, {IJCAI} 2024, Jeju, South Korea, August 3-9, 2024},
  pages = {8299--8307},
  publisher = {ijcai.org},
  year = {2024},
  url = {https://www.ijcai.org/proceedings/2024/917},
  bibsource = {dblp computer science bibliography, https://dblp.org}
}

@inproceedings{DBLP:conf/emnlp/ManakulLG23,
  author = {Manakul Potsawee and Adian Liusie and Mark J. F. Gales},
  editor = {Houda Bouamor and Juan Pino and Kalika Bali},
  title = {SelfCheckGPT: Zero-Resource Black-Box Hallucination Detection for Generative Large Language Models},
  booktitle = {Proceedings of the 2023 Conference on Empirical Methods in Natural Language Processing, {EMNLP} 2023, Singapore, December 6-10, 2023},
  pages = {9004--9017},
  publisher = {Association for Computational Linguistics},
  year = {2023},
  url = {https://doi.org/10.18653/v1/2023.emnlp-main.557},
  doi = {10.18653/V1/2023.EMNLP-MAIN.557},
  bibsource = {dblp computer science bibliography, https://dblp.org}
}

@article{DBLP:journals/corr/abs-2410-02899,
  author = {Deema Alnuhait and Neeraja Kirtane and Muhammad Khalifa and Hao Peng},
  title = {FactCheckmate: Preemptively Detecting and Mitigating Hallucinations in LMs},
  journal = {CoRR},
  volume = {abs/2410.02899},
  year = {2024},
  url = {https://doi.org/10.48550/arXiv.2410.02899},
  doi = {10.48550/ARXIV.2410.02899},
  eprinttype = {arXiv},
  eprint = {2410.02899},
  bibsource = {dblp computer science bibliography, https://dblp.org}
}

@inproceedings{DBLP:conf/emnlp/0038GLZPK24,
  author = {Qing Li and Jiahui Geng and Chenyang Lyu and Derui Zhu and Maxim Panov and Fakhri Karray},
  editor = {Yaser Al{-}Onaizan and Mohit Bansal and Yun{-}Nung Chen},
  title = {Reference-free Hallucination Detection for Large Vision-Language Models},
  booktitle = {Findings of the Association for Computational Linguistics: {EMNLP} 2024, Miami, Florida, USA, November 12-16, 2024},
  pages = {4542--4551},
  publisher = {Association for Computational Linguistics},
  year = {2024},
  url = {https://aclanthology.org/2024.findings-emnlp.262},
  bibsource = {dblp computer science bibliography, https://dblp.org}
}

@article{hallucinations-leaderboard,
  author = {Giwon Hong and Aryo Pradipta Gema and Rohit Saxena and Xiaotang Du and Ping Nie and Yu Zhao and Laura Perez{-}Beltrachini and Max Ryabinin and Xuanli He and Cl{\'{e}}mentine Fourrier and Pasquale Minervini},
  title = {The Hallucinations Leaderboard - An Open Effort to Measure Hallucinations in Large Language Models},
  journal = {CoRR},
  volume = {abs/2404.05904},
  year = {2024},
  url = {https://doi.org/10.48550/arXiv.2404.05904},
  doi = {10.48550/ARXIV.2404.05904},
  eprinttype = {arXiv},
  eprint = {2404.05904},
  bibsource = {dblp computer science bibliography, https://dblp.org}
}

@inproceedings{DBLP:conf/ijcai/0016SGW000LZC24,
  author = {Xiang Chen and Duanzheng Song and Honghao Gui and Chenxi Wang and Ningyu Zhang and Yong Jiang and Fei Huang and Chengfei Lyu and Dan Zhang and Huajun Chen},
  title = {FactCHD: Benchmarking Fact-Conflicting Hallucination Detection},
  booktitle = {Proceedings of the Thirty-Third International Joint Conference on Artificial Intelligence, {IJCAI} 2024, Jeju, South Korea, August 3-9, 2024},
  pages = {6216--6224},
  publisher = {ijcai.org},
  year = {2024},
  url = {https://www.ijcai.org/proceedings/2024/687},
  bibsource = {dblp computer science bibliography, https://dblp.org}
}

@inproceedings{DBLP:conf/acl/ZhangPTZJSMM24,
  author = {Xiaoying Zhang and Baolin Peng and Ye Tian and Jingyan Zhou and Lifeng Jin and Linfeng Song and Haitao Mi and Helen Meng},
  editor = {Lun{-}Wei Ku and Andre Martins and Vivek Srikumar},
  title = {Self-Alignment for Factuality: Mitigating Hallucinations in LLMs via Self-Evaluation},
  booktitle = {Proceedings of the 62nd Annual Meeting of the Association for Computational Linguistics (Volume 1: Long Papers), {ACL} 2024, Bangkok, Thailand, August 11-16, 2024},
  pages = {1946--1965},
  publisher = {Association for Computational Linguistics},
  year = {2024},
  url = {https://doi.org/10.18653/v1/2024.acl-long.107},
  doi = {10.18653/V1/2024.ACL-LONG.107},
  bibsource = {dblp computer science bibliography, https://dblp.org}
}

@inproceedings{DBLP:conf/www/ShenLFRBN23,
  author = {Jiaming Shen and Jialu Liu and Daniel Finnie and Negar Rahmati and Mike Bendersky and Marc Najork},
  editor = {Ying Ding and Jie Tang and Juan F. Sequeda and Lora Aroyo and Carlos Castillo and Geert{-}Jan Houben},
  title = {"Why is this misleading?": Detecting News Headline Hallucinations with Explanations},
  booktitle = {Proceedings of the {ACM} Web Conference 2023, {WWW} 2023, Austin, TX, USA, 30 April 2023 - 4 May 2023},
  pages = {1662--1672},
  publisher = {{ACM}},
  year = {2023},
  url = {https://doi.org/10.1145/3543507.3583375},
  doi = {10.1145/3543507.3583375},
  bibsource = {dblp computer science bibliography, https://dblp.org}
}

@inproceedings{DBLP:conf/naacl/PagnoniBT21,
  author = {Artidoro Pagnoni and Vidhisha Balachandran and Yulia Tsvetkov},
  editor = {Kristina Toutanova and Anna Rumshisky and Luke Zettlemoyer and Dilek Hakkani{-}T{\"{u}}r and Iz Beltagy and Steven Bethard and Ryan Cotterell and Tanmoy Chakraborty and Yichao Zhou},
  title = {Understanding Factuality in Abstractive Summarization with {FRANK:} {A} Benchmark for Factuality Metrics},
  booktitle = {Proceedings of the 2021 Conference of the North American Chapter of the Association for Computational Linguistics: Human Language Technologies, {NAACL-HLT} 2021, Online, June 6-11, 2021},
  pages = {4812--4829},
  publisher = {Association for Computational Linguistics},
  year = {2021},
  url = {https://doi.org/10.18653/v1/2021.naacl-main.383},
  doi = {10.18653/V1/2021.NAACL-MAIN.383},
  bibsource = {dblp computer science bibliography, https://dblp.org}
}

@inproceedings{DBLP:conf/emnlp/DasSS22,
  author = {Souvik Das and Sougata Saha and Rohini K. Srihari},
  editor = {Yoav Goldberg and Zornitsa Kozareva and Yue Zhang},
  title = {Diving Deep into Modes of Fact Hallucinations in Dialogue Systems},
  booktitle = {Findings of the Association for Computational Linguistics: {EMNLP} 2022, Abu Dhabi, United Arab Emirates, December 7-11, 2022},
  pages = {684--699},
  publisher = {Association for Computational Linguistics},
  year = {2022},
  url = {https://doi.org/10.18653/v1/2022.findings-emnlp.48},
  doi = {10.18653/V1/2022.FINDINGS-EMNLP.48},
  bibsource = {dblp computer science bibliography, https://dblp.org}
}

@inproceedings{DBLP:conf/acl/MallenAZDKH23,
  author = {Alex Mallen and Akari Asai and Victor Zhong and Rajarshi Das and Daniel Khashabi and Hannaneh Hajishirzi},
  editor = {Anna Rogers and Jordan L. Boyd{-}Graber and Naoaki Okazaki},
  title = {When Not to Trust Language Models: Investigating Effectiveness of Parametric and Non-Parametric Memories},
  booktitle = {Proceedings of the 61st Annual Meeting of the Association for Computational Linguistics (Volume 1: Long Papers), {ACL} 2023, Toronto, Canada, July 9-14, 2023},
  pages = {9802--9822},
  publisher = {Association for Computational Linguistics},
  year = {2023},
  url = {https://doi.org/10.18653/v1/2023.acl-long.546},
  doi = {10.18653/V1/2023.ACL-LONG.546},
  bibsource = {dblp computer science bibliography, https://dblp.org}
}

@article{rabby2024mc,
  title   = {MC-NEST--Enhancing Mathematical Reasoning in Large Language Models with a Monte Carlo Nash Equilibrium Self-Refine Tree},
  author  = {Rabby, Gollam and Keya, Farhana and Zamil, Parvez and Auer, S{\"o}ren},
  journal = {arXiv preprint arXiv:2411.15645},
  year    = {2024},
  url     = {https://doi.org/10.48550/arXiv.2411.15645},
  doi     = {10.48550/arXiv.2411.15645},
  eprinttype = {arXiv},
  eprint  = {2411.15645}
}

@inproceedings{DBLP:conf/acl/StranisciDMPRC23,
  author = {Marco Antonio Stranisci and Rossana Damiano and Enrico Mensa and Viviana Patti and Daniele Paolo Radicioni and Tommaso Caselli},
  editor = {Anna Rogers and Jordan L. Boyd{-}Graber and Naoaki Okazaki},
  title = {WikiBio: a Semantic Resource for the Intersectional Analysis of Biographical Events},
  booktitle = {Proceedings of the 61st Annual Meeting of the Association for Computational Linguistics (Volume 1: Long Papers), {ACL} 2023, Toronto, Canada, July 9-14, 2023},
  pages = {12370--12384},
  publisher = {Association for Computational Linguistics},
  year = {2023},
  url = {https://doi.org/10.18653/v1/2023.acl-long.691},
  doi = {10.18653/V1/2023.ACL-LONG.691},
  bibsource = {dblp computer science bibliography, https://dblp.org}
}

@inproceedings{DBLP:conf/naacl/ThorneVCM18,
  author = {James Thorne and Andreas Vlachos and Christos Christodoulopoulos and Arpit Mittal},
  editor = {Marilyn A. Walker and Heng Ji and Amanda Stent},
  title = {{FEVER:} a Large-scale Dataset for Fact Extraction and VERification},
  booktitle = {Proceedings of the 2018 Conference of the North American Chapter of the Association for Computational Linguistics: Human Language Technologies, {NAACL-HLT} 2018, New Orleans, Louisiana, USA, June 1-6, 2018, Volume 1 (Long Papers)},
  pages = {809--819},
  publisher = {Association for Computational Linguistics},
  year = {2018},
  url = {https://doi.org/10.18653/v1/n18-1074},
  doi = {10.18653/V1/N18-1074},
  bibsource = {dblp computer science bibliography, https://dblp.org}
}

@article{DBLP:journals/corr/abs-2303-08774,
  author = {OpenAI},
  title = {{GPT-4} Technical Report},
  journal = {CoRR},
  volume = {abs/2303.08774},
  year = {2023},
  url = {https://doi.org/10.48550/arXiv.2303.08774},
  doi = {10.48550/ARXIV.2303.08774},
  eprinttype = {arXiv},
  eprint = {2303.08774},
  bibsource = {dblp computer science bibliography, https://dblp.org}
}

@inproceedings{DBLP:conf/acl/LinHE22,
  author = {Stephanie Lin and Jacob Hilton and Owain Evans},
  editor = {Smaranda Muresan and Preslav Nakov and Aline Villavicencio},
  title = {TruthfulQA: Measuring How Models Mimic Human Falsehoods},
  booktitle = {Proceedings of the 60th Annual Meeting of the Association for Computational Linguistics (Volume 1: Long Papers), {ACL} 2022, Dublin, Ireland, May 22-27, 2022},
  pages = {3214--3252},
  publisher = {Association for Computational Linguistics},
  year = {2022},
  url = {https://doi.org/10.18653/v1/2022.acl-long.229},
  doi = {10.18653/V1/2022.ACL-LONG.229},
  bibsource = {dblp computer science bibliography, https://dblp.org}
}

@inproceedings{DBLP:conf/emnlp/LiCZNW23,
  author = {Junyi Li and Xiaoxue Cheng and Xin Zhao and Jian{-}Yun Nie and Ji{-}Rong Wen},
  editor = {Houda Bouamor and Juan Pino and Kalika Bali},
  title = {HaluEval: {A} Large-Scale Hallucination Evaluation Benchmark for Large Language Models},
  booktitle = {Proceedings of the 2023 Conference on Empirical Methods in Natural Language Processing, {EMNLP} 2023, Singapore, December 6-10, 2023},
  pages = {6449--6464},
  publisher = {Association for Computational Linguistics},
  year = {2023},
  url = {https://doi.org/10.18653/v1/2023.emnlp-main.397},
  doi = {10.18653/V1/2023.EMNLP-MAIN.397},
  bibsource = {dblp computer science bibliography, https://dblp.org}
}

@inproceedings{DBLP:conf/acl/LiuZBMSCD22,
  author = {Tianyu Liu and Yizhe Zhang and Chris Brockett and Yi Mao and Zhifang Sui and Weizhu Chen and Bill Dolan},
  editor = {Smaranda Muresan and Preslav Nakov and Aline Villavicencio},
  title = {A Token-level Reference-free Hallucination Detection Benchmark for Free-form Text Generation},
  booktitle = {Proceedings of the 60th Annual Meeting of the Association for Computational Linguistics (Volume 1: Long Papers), {ACL} 2022, Dublin, Ireland, May 22-27, 2022},
  pages = {6723--6737},
  publisher = {Association for Computational Linguistics},
  year = {2022},
  url = {https://doi.org/10.18653/v1/2022.acl-long.464},
  doi = {10.18653/V1/2022.ACL-LONG.464},
  bibsource = {dblp computer science bibliography, https://dblp.org}
}

@inproceedings{DBLP:conf/nips/DettmersPHZ23,
  author    = {Tim Dettmers and Artidoro Pagnoni and Ari Holtzman and Luke Zettlemoyer},
  editor    = {Alice Oh and Tristan Naumann and Amir Globerson and Kate Saenko and Moritz Hardt and Sergey Levine},
  title     = {QLoRA: Efficient Finetuning of Quantized LLMs},
  booktitle = {Advances in Neural Information Processing Systems 36 (NeurIPS 2023), New Orleans, LA, USA, December 10--16, 2023},
  year      = {2023},
  url       = {https://doi.org/10.48550/arXiv.2305.14314},
  doi       = {10.48550/arXiv.2305.14314},
  eprinttype = {arXiv},
  eprint    = {2305.14314},
  bibsource = {dblp computer science bibliography, https://dblp.org}
}

@inproceedings{DBLP:conf/naacl/WilliamsNB18,
  author = {Adina Williams and Nikita Nangia and Samuel R. Bowman},
  editor = {Marilyn A. Walker and Heng Ji and Amanda Stent},
  title = {A Broad-Coverage Challenge Corpus for Sentence Understanding through Inference},
  booktitle = {Proceedings of the 2018 Conference of the North American Chapter of the Association for Computational Linguistics: Human Language Technologies, {NAACL-HLT} 2018, New Orleans, Louisiana, USA, June 1-6, 2018, Volume 1 (Long Papers)},
  pages = {1112--1122},
  publisher = {Association for Computational Linguistics},
  year = {2018},
  url = {https://doi.org/10.18653/v1/n18-1101},
  doi = {10.18653/V1/N18-1101},
  bibsource = {dblp computer science bibliography, https://dblp.org}
}

@inproceedings{DBLP:conf/emnlp/Peng0S24,
  author = {Letian Peng and Zilong Wang and Jingbo Shang},
  editor = {Yaser Al{-}Onaizan and Mohit Bansal and Yun{-}Nung Chen},
  title = {Incubating Text Classifiers Following User Instruction with Nothing but {LLM}},
  booktitle = {Proceedings of the 2024 Conference on Empirical Methods in Natural Language Processing, {EMNLP} 2024, Miami, FL, USA, November 12-16, 2024},
  pages = {3753--3766},
  publisher = {Association for Computational Linguistics},
  year = {2024},
  url = {https://aclanthology.org/2024.emnlp-main.220},
  bibsource = {dblp computer science bibliography, https://dblp.org}
}

@inproceedings{DBLP:journals/corr/abs-1301-3781,
  author = {Tom{\'{a}}s Mikolov and Kai Chen and Greg Corrado and Jeffrey Dean},
  editor = {Yoshua Bengio and Yann LeCun},
  title = {Efficient Estimation of Word Representations in Vector Space},
  booktitle = {1st International Conference on Learning Representations, {ICLR} 2013, Scottsdale, Arizona, USA, May 2-4, 2013, Workshop Track Proceedings},
  year = {2013},
  url = {http://arxiv.org/abs/1301.3781},
  bibsource = {dblp computer science bibliography, https://dblp.org}
}

@article{DBLP:journals/corr/abs-2303-10420,
  author = {Junjie Ye and Xuanting Chen and Nuo Xu and Can Zu and Zekai Shao and Shichun Liu and Yuhan Cui and Zeyang Zhou and Chao Gong and Yang Shen and Jie Zhou and Siming Chen and Tao Gui and Qi Zhang and Xuanjing Huang},
  title = {A Comprehensive Capability Analysis of {GPT-3} and {GPT-3.5} Series Models},
  journal = {CoRR},
  volume = {abs/2303.10420},
  year = {2023},
  url = {https://doi.org/10.48550/arXiv.2303.10420},
  doi = {10.48550/ARXIV.2303.10420},
  eprinttype = {arXiv},
  eprint = {2303.10420},
  bibsource = {dblp computer science bibliography, https://dblp.org}
}

@article{abdin2024phi,
  title       = {Phi-3 Technical Report: A Highly Capable Language Model Locally on Your Phone},
  author      = {Abdin, Marah and Aneja, Jyoti and Awadalla, Hany and Awadallah, Ahmed and Awan, Ammar Ahmad and Bach, Nguyen and Bahree, Amit and Bakhtiari, Arash and Bao, Jianmin and Behl, Harkirat and others},
  journal     = {arXiv preprint arXiv:2404.14219},
  year        = {2024},
  url         = {https://doi.org/10.48550/arXiv.2404.14219},
  doi         = {10.48550/arXiv.2404.14219},
  eprinttype  = {arXiv},
  eprint      = {2404.14219}
}

@article{dubey2024llama,
  title       = {The Llama 3 Herd of Models},
  author      = {Dubey, Abhimanyu and Jauhri, Abhinav and Pandey, Abhinav and Kadian, Abhishek and Al-Dahle, Ahmad and Letman, Aiesha and Mathur, Akhil and Schelten, Alan and Yang, Amy and Fan, Angela and others},
  journal     = {arXiv preprint arXiv:2407.21783},
  year        = {2024},
  url         = {https://doi.org/10.48550/arXiv.2407.21783},
  doi         = {10.48550/arXiv.2407.21783},
  eprinttype  = {arXiv},
  eprint      = {2407.21783}
}

@article{jiang2023mistral,
  title       = {Mistral 7B},
  author      = {Jiang, Albert Q and Sablayrolles, Alexandre and Mensch, Arthur and Bamford, Chris and Chaplot, Devendra Singh and Casas, Diego de las and Bressand, Florian and Lengyel, Gianna and Lample, Guillaume and Saulnier, Lucile and others},
  journal     = {arXiv preprint arXiv:2310.06825},
  year        = {2023},
  url         = {https://doi.org/10.48550/arXiv.2310.06825},
  doi         = {10.48550/arXiv.2310.06825},
  eprinttype  = {arXiv},
  eprint      = {2310.06825}
}

@article{team2024gemma,
  title = {Gemma: Open models based on gemini research and technology},
  author = {Team, Gemma and Mesnard, Thomas and Hardin, Cassidy and Dadashi, Robert and Bhupatiraju, Surya and Pathak, Shreya and Sifre, Laurent and Rivi{\`e}re, Morgane and Kale, Mihir Sanjay and Love, Juliette and others},
  journal = {arXiv preprint arXiv:2403.08295},
  year = {2024} }

@article{logiqa,
  title     = {LogiQA 2.0—An Improved Dataset for Logical Reasoning in Natural Language Understanding},
  author    = {Liu, Hanmeng and Liu, Jian and Cui, Leyang and Teng, Zhiyang and Duan, Nan and Zhou, Ming and Zhang, Yue},
  journal   = {IEEE/ACM Transactions on Audio, Speech, and Language Processing},
  year      = {2023},
  volume    = {31},
  pages     = {2947--2962},
  doi       = {10.1109/TASLP.2023.3293046},
  url       = {https://doi.org/10.1109/TASLP.2023.3293046}
}

@article{fgprm2024,
  title       = {FG-PRM: Fine-grained Hallucination Detection and Mitigation in Language Model Mathematical Reasoning},
  author      = {Li, Ruosen and Luo, Yiming and Du, Xinya},
  journal     = {arXiv preprint arXiv:2410.06304},
  year        = {2024},
  url         = {https://doi.org/10.48550/arXiv.2410.06304},
  doi         = {10.48550/arXiv.2410.06304},
  eprinttype  = {arXiv},
  eprint      = {2410.06304}
}

@article{rewardmodel-openai,
  author      = {Ziegler, Daniel M. and Stiennon, Nisan and Wu, Jeffrey and Brown, Tom B. and Radford, Alec and Amodei, Dario and Christiano, Paul and Irving, Geoffrey},
  title       = {Fine-Tuning Language Models from Human Preferences},
  journal     = {CoRR},
  volume      = {abs/1909.08593},
  year        = {2019},
  url         = {https://doi.org/10.48550/arXiv.1909.08593},
  doi         = {10.48550/arXiv.1909.08593},
  eprinttype  = {arXiv},
  eprint      = {1909.08593}
}

@article{rewardmath,
  author      = {Kim, Sunghwan and Kang, Dongjin and Kwon, Taeyoon and Chae, Hyungjoo and Won, Jungsoo and Lee, Dongha and Yeo, Jinyoung},
  title       = {Evaluating Robustness of Reward Models for Mathematical Reasoning},
  journal     = {CoRR},
  volume      = {abs/2410.01729},
  year        = {2024},
  url         = {https://doi.org/10.48550/arXiv.2410.01729},
  doi         = {10.48550/arXiv.2410.01729},
  eprinttype  = {arXiv},
  eprint      = {2410.01729}
}

@article{lightman2023letsverify,
  author      = {Lightman, Hunter and Kosaraju, Vineet and Burda, Yura and Edwards, Harri and Baker, Bowen and Lee, Teddy and Leike, Jan and Schulman, John and Sutskever, Ilya and Cobbe, Karl},
  title       = {Let's Verify Step by Step},
  journal     = {CoRR},
  volume      = {abs/2305.20050},
  year        = {2023},
  url         = {https://doi.org/10.48550/arXiv.2305.20050},
  doi         = {10.48550/arXiv.2305.20050},
  eprinttype  = {arXiv},
  eprint      = {2305.20050}
}

@inproceedings{wang2024mathshepherd,
  title       = {Math-Shepherd: Verify and Reinforce {LLMs} Step-by-step without Human Annotations},
  author      = {Wang, Peiyi and Li, Lei and Shao, Zhihong and Xu, Runxin and Dai, Damai and Li, Yifei and Chen, Deli and Wu, Yu and Sui, Zhifang},
  booktitle   = {Proceedings of the 62nd Annual Meeting of the Association for Computational Linguistics (Volume 1: Long Papers)},
  year        = {2024},
  pages       = {9426--9439},
  address     = {Bangkok, Thailand},
  publisher   = {Association for Computational Linguistics},
  url         = {https://aclanthology.org/2024.acl-long.510},
  doi         = {10.18653/v1/2024.acl-long.510}
}

@article{uesato2022process,
  title = {Solving Math Word Problems with Process- and Outcome-based Feedback},
  author = {Uesato, Jonathan and Kushman, Nate and Kumar, Ramana and Song, Francis and Siegel, Noah and Wang, Lisa and Creswell, Antonia and Irving, Geoffrey and Higgins, Irina},
  journal = {arXiv preprint arXiv:2211.14275},
  year = {2022},
  url = {https://arxiv.org/abs/2211.14275}
}

@article{farquhar2024,
  author    = {Farquhar, Sebastian and Kossen, Jannik and Kuhn, Lorenz and Gal, Yarin},
  title     = {Detecting hallucinations in large language models using semantic entropy},
  journal   = {Nature},
  year      = {2024},
  volume    = {627},
  number    = {8002},
  pages     = {768--773},
  publisher = {Springer Nature},
  doi       = {10.1038/s41586-024-07421-0},
  url       = {https://doi.org/10.1038/s41586-024-07421-0}
}

@article{gsm8k,
  author = "Karl Cobbe and Vineet Kosaraju and Mohammad Bavarian and Mark Chen and Heewoo Jun and Lukasz Kaiser and Matthias Plappert and Jerry Tworek and Jacob Hilton and Reiichiro Nakano and Christopher Hesse and John Schulman",
  title = "{Training Verifiers to Solve Math Word Problems}",
  journal = "{arXiv preprint arXiv:2110.14168}",
  year = "2021",
  url = {https://arxiv.org/abs/2110.14168}
}

@article{math,
  author = {Karl Cobbe and Vineet Kosaraju and Mohammad Bavarian and Mark Chen and Heewoo Jun and Lukasz Kaiser and Matthias Plappert and Jerry Tworek and Jacob Hilton and Reiichiro Nakano and Christopher Hesse and John Schulman},
  title = "{Measuring Mathematical Problem Solving With the MATH Dataset}",
  journal = "{arXiv preprint arXiv:2103.03874}",
  year = "2021",
  url = {https://arxiv.org/abs/2103.03874}
}

@inproceedings{svamp,
  title = "Are {NLP} Models really able to Solve Simple Math Word Problems?",
  author = "Patel, Arkil and Bhattamishra, Satwik and Goyal, Navin",
  editor = "Toutanova, Kristina and Rumshisky, Anna and Zettlemoyer, Luke and Hakkani-Tur, Dilek and Beltagy, Iz and Bethard, Steven and Cotterell, Ryan and Chakraborty, Tanmoy and Zhou, Yichao",
  booktitle = "Proceedings of the 2021 Conference of the North American Chapter of the Association for Computational Linguistics: Human Language Technologies",
  month = jun,
  year = "2021",
  address = "Online",
  publisher = "Association for Computational Linguistics",
  url = "https://aclanthology.org/2021.naacl-main.168/",
  doi = "10.18653/v1/2021.naacl-main.168",
  pages = "2080--2094"
}

@article{DBLP:journals/corr/abs-1907-11692,
  author    = {Yinhan Liu and
               Myle Ott and
               Naman Goyal and
               Jingfei Du and
               Mandar Joshi and
               Danqi Chen and
               Omer Levy and
               Mike Lewis and
               Luke Zettlemoyer and
               Veselin Stoyanov},
  title     = {RoBERTa: {A} Robustly Optimized {BERT} Pretraining Approach},
  journal   = {CoRR},
  volume    = {abs/1907.11692},
  year      = {2019},
  url       = {http://arxiv.org/abs/1907.11692},
  archivePrefix = {arXiv},
  eprint    = {1907.11692},
  bibsource = {dblp computer science bibliography, https://dblp.org}
}

\bigskip


\newpage

\section{Appendix}








\subsection{Prompts in Experiment}

\textbf{Specialized Detection \textcolor{black}{module} on Pretrained LLMs}

\begin{tcolorbox}[colback=gray!5!white, colframe=gray!75!black, title=Zero-shot prompt]
\textbf{Statement 1:} \texttt{\{sentence\}} \\[6pt]
\textbf{Statement 2:} \texttt{\{sample\}} \\[6pt]
\textbf{Task:} Analyze if these statements contradict or agree. \\[6pt]
\textbf{Instructions:} Answer with \texttt{contradict} or \texttt{agree}.
\end{tcolorbox}

\begin{tcolorbox}[colback=gray!5!white, colframe=gray!75!black, title=Chain-of-thought prompt]
\textbf{Statement 1:} \texttt{\{sentence\}} \\[6pt]
\textbf{Statement 2:} \texttt{\{sample\}} \\[6pt]
\textbf{Task:} Let's reason step by step to see if these statements are related. \\[6pt]
\textbf{Instructions:} Consider whether the second statement logically follows from or contradicts the first one.  
Answer with \texttt{contradict} or \texttt{entailment}.
\end{tcolorbox}

\textbf{Contextual Consistency \textcolor{black}{module}}

\begin{tcolorbox}[colback=gray!5!white, colframe=gray!75!black, title=Zero-shot prompt]
\textbf{Context:} \texttt{\{context\}} \\[6pt]
\textbf{Sentence:} \texttt{\{sentence\}} \\[6pt]
\textbf{Task:} Is the sentence supported by the context above? \\[6pt]
\textbf{Instructions:} Answer \texttt{Yes} or \texttt{No}. \\[6pt]
\textbf{Answer:}
\end{tcolorbox}

\begin{tcolorbox}[colback=gray!5!white, colframe=gray!75!black, title=Chain-of-thought prompt]
\textbf{Context:} \texttt{\{context\}} \\[6pt]
\textbf{Sentence:} \texttt{\{sentence\}} \\[6pt]
\textbf{Task:} Check if the sentence is supported by the context. \\[6pt]
\textbf{Instructions:} Think step by step:  
1. Understand what the sentence claims.  
2. Check if the context provides evidence for this claim.  
3. Decide if the sentence is supported by the context.  

After reasoning, respond with ONLY one word: \texttt{Yes} or \texttt{No}. \\[6pt]
\textbf{Answer:}
\end{tcolorbox}






\subsection{AIME Hallucination Detection Dataset}

This dataset is designed for detecting hallucinations in LLMs, particularly for complex mathematical problems. It can be used for evaluation and related research. The dataset contains the LLM's initial response and sampled response, along with their corresponding labels. Column headers and a description of the dataset are provided \textcolor{black}{in Table~\ref{tab:aime-columns}}.

\begin{table}[h!]
\centering
\begin{tabular}{@{}p{4cm}p{10cm}@{}}
\toprule
\textbf{Column Name} & \textbf{Description} \\
\midrule
Year & Year in which the problem appeared in AIME. \\
Set & The set in which the problem appeared. \\
Problem Number & Problem number. \\
URL & URL for the problem. \\
Problem Statement & The mathematical problem/query. \\
Exact Answer & The exact final answer to the problem. \\
Solution 1--13 & Solutions provided by human experts. \\
LLM Solution (gpt-4o) & LLM's generated solution. \\
Exact Answer (gpt-4o) & Exact answer of LLM's output. \\
Annotation & \begin{tabular}[t]{@{}l@{}}
0: Accurate \\ 
1: Minor\_inaccurate \\ 
2: Major\_inaccurate
\end{tabular} \\
Sampled Responses & LLM's sampled outputs. \\
\bottomrule
\end{tabular}
\caption{AIME columns and descriptions}
\label{tab:aime-columns}
\end{table}

\subsection{WikiBio Hallucination Detection Dataset}

This dataset includes Wikipedia-like passages generated by GPT-3 (text-davinci-003) with human annotations for sentence-level accuracy. It is used to evaluate hallucination detection in LLMs. Column headers and descriptions are provided \textcolor{black}{in Table~\ref{tab:wikibio-columns}}.

\begin{table}[h!]
\centering
\begin{tabular}{@{}p{4cm}p{10cm}@{}}
\toprule
\textbf{Column Name} & \textbf{Description} \\
\midrule
gpt3\_text & GPT-3 generated passage. \\
wiki\_bio\_text & Actual Wikipedia passage (first paragraph)\\
gpt3\_sentences & gpt3\_text split into sentences using spacy\\
wiki\_bio\_test\_idx & ID of the concept/individual from the original wikibio dataset (testset) \\
Annotation & \begin{tabular}[t]{@{}l@{}l@{}}
human annotation at the sentence level\\
Accurate \\ 
Minor\_inaccurate \\ 
Major\_inaccurate
\end{tabular} \\
gpt3\_text\_samples & list of 20 sampled passages \\
\bottomrule
\end{tabular}
\caption{WikiBio columns and descriptions.}
\label{tab:wikibio-columns}
\end{table}








\newpage

\subsection{Additional Datasets}
\subsubsection{Dataset LogicQA}
\label{appendix:logicqa}
The NLI version of the LogiQA~2.0 dataset consists of 39k premise–hypothesis pairs annotated as 
\textit{Entailed} or \textit{Not Entailed}. 
Unlike standard NLI datasets, LogiQA~2.0 emphasizes logical inference beyond the sentence level, 
making it suitable for evaluating transfer to multi-step reasoning tasks in AIME-MATH.  An example instance from LogiQA2NLI is shown \textcolor{black}{in Listing~\ref{lst:logiqa-example}.} 
\begin{lstlisting}[style=jsonstyle,
    caption={Example instance from the LogiQA2NLI dataset.},
    label={lst:logiqa-example}]
{
  "label": "not entailed",
  "major_premise": [
    "A, B, and C have three balls, one is red, one is blue, and the other is yellow"
  ],
  "conclusion": "A is red, B is blue, C is yellow",
  "minor_premise": "C is bigger than the yellow ball, A and the blue ball are not the same size, and the blue ball is smaller than C"
}
\end{lstlisting}

In our experiments, we fine-tuned several LLMs on the QA2NLI from LogiQA~2.0. 
The selected LLMs include:
\begin{itemize}
    \item \textbf{LLaMA3-8B}, as a general-purpose baseline;
    \item \textbf{Qwen2.5-7B}, chosen for its instruction-following and long-context reasoning capabilities;
    \item \textbf{Qwen2.5-Math-7B}, with prior fine-tuning on mathematical chain-of-thought supervision;
    \item \textbf{Phi-3}, based on its strong performance in earlier detection experiments.
\end{itemize}

To ensure comparability, all LLMs were fine-tuned under identical hyperparameter settings: 6 epochs, a learning rate of $1 \times 10^{-5}$, and the Adam optimizer. This configuration was found to be the most effective in our experimental setup.

\subsubsection{Dataset RewardMATH}
\label{appendix:rewardmath}
The RewardMATH dataset consists of 4,830 examples derived from 483 problems. 
Each problem is paired with one correct and multiple incorrect solutions, 
spanning seven domains: Algebra, Intermediate Algebra, Prealgebra, 
Number Theory, Precalculus, Geometry, and Counting \& Probability. The following excerpts \textcolor{black}{(Listing~\ref{lst:rewardmath-chosen} and ~\ref{lst:rewardmath-rejected})} illustrate the structure of RewardMATH examples (simplified). We report both a \texttt{chosen} (preferred) and a \texttt{rejected} (non-preferred) solution.

\begin{lstlisting}[style=jsonstyle,
    caption={RewardMATH example with a chosen solution.},
    label={lst:rewardmath-chosen}]
{
  "problem": "$441+2(21)(19)+361=x$. Solve for $x$.",
  "eval_solution": [
    "Simplify: 2(21)(19) = 798.",
    "Equation becomes 441 + 798 + 361 = x.",
    "This equals (21+19)^2 = 1600.",
    "Therefore, x = 1600."
  ],
  "solution_type": "chosen",
  "level": "Algebra"
}
\end{lstlisting}

\begin{lstlisting}[style=jsonstyle,
    caption={RewardMATH example with a rejected solution.},
    label={lst:rewardmath-rejected}]
{
  "problem": "$441+2(21)(19)+361=x$. Solve for $x$.",
  "eval_solution": [
    "Calculate 2(21)(19) = 800.",
    "Equation becomes 441 + 800 + 361 = 1602.",
    "Therefore, x = 1602."
  ],
  "solution_type": "rejected",
  "level": "Algebra"
}
\end{lstlisting}

All reward models were trained using the Hugging Face TRL RewardTrainer\footnote{\url{https://huggingface.co/docs/trl/main/en/reward_trainer}}, for three epochs with a learning rate of $1 \times 10^{-5}$.

\newpage
\subsubsection{Dataset MathShepherd}
\label{appendix:mathshepherd}

The MathShepherd dataset~\cite{wang2024mathshepherd} provides large-scale process-level supervision 
by automatically generating step-level labels. Each intermediate step in a reasoning trace is scored 
based on whether it can lead to the correct solution. The dataset contains 445k instances 
(422k train and 22.2k test), enabling scalable process supervision without requiring manual annotation. An example from MathShepherd is reported \textcolor{black}{in Listing~\ref{lst:mathshepherd-example}.}

\begin{lstlisting}[style=jsonstyle,
    caption={Example instance from the MathShepherd dataset.},
    label={lst:mathshepherd-example}]
{
  "problem": "If two distinct numbers are randomly chosen from the set {1, 2, 3, 4, 5}, 
  what is the probability that the smaller number is a divisor of the larger number? 
  Express your answer as a common fraction.",
  "steps": [
    "Step 1: There are a total of C(5,2)=10 pairs of numbers that can be chosen.",
    "Step 2: List them all: (1,2), (1,3), (1,4), (1,5), (2,3), (2,4), (2,5), (3,4), (3,5), (4,5).",
    "Step 3: Valid pairs are (1,2), (1,3), (1,4), (1,5), (2,4).",
    "Step 4: So, there are 5 pairs that satisfy the condition.",
    "Step 5: Probability = 5/10 = 1/2."
  ],
  "labels": [true, true, true, true, true]
}
\end{lstlisting}

LLMs were fine-tuned on 30\% of the official training split, as preliminary experiments showed 
that using the full dataset provided no additional improvement. We fine-tuned for three epochs, 
with a learning rate of $1 \times 10^{-5}$, the Adam optimizer, and a fixed random seed of 10, 
using the Hugging Face TRL \texttt{PRMTrainer}\footnote{\url{https://huggingface.co/docs/trl/main/prm_trainer}}.

\newpage
\subsubsection{Dataset PRM800K}
\label{appendix:prm800k}

The PRM800K dataset provides step-level supervision: each intermediate reasoning step is annotated 
as \texttt{true} (correct) or \texttt{false} (incorrect). 
An example instance is shown \textcolor{black}{in Listing~\ref{lst:prm800k-instance}.}

\begin{lstlisting}[style=jsonstyle,
    caption={Example instance from the PRM800K dataset.},
    label={lst:prm800k-instance}]
{
  "prompt": "How many seconds are in 7.8 minutes?",
  "completions": [
    "7.8 minutes is the same as 7 minutes and 0.8 minutes.",
    "Right, and since there are 60 seconds in a minute, then there are 60 * 7 = 420 seconds in 7 minutes.",
    "And since there are 60 seconds in a minute, then there are 60 * 0.8 = 48 seconds in 0.8 minutes.",
    "So, in total, there are 420 + 48 = 468 seconds in 7.8 minutes.",
    "Right. Let's check our work. 7.8 minutes is the same as 7 minutes and 0.8 minutes."
  ],
  "labels": [true, true, true, true, false]
}
\end{lstlisting}

All LLMa were fine-tuned using the Hugging Face TRL \texttt{PRMTrainer}\footnote{\url{https://huggingface.co/docs/trl/main/prm_trainer}} 
for three epochs, with a learning rate of $1 \times 10^{-5}$, the Adam optimizer, and a fixed random seed of 10.  

\section{Real-World Application Case Study}

\renewcommand{\arraystretch}{1.3}
\begin{table*}[htbp]
\centering
\small 
\begin{adjustbox}{center,max width=\textwidth}
\begin{tabular}{p{2.8cm}|p{3.2cm}|p{3.2cm}|
                >{\centering\arraybackslash}p{1.8cm}|
                >{\centering\arraybackslash}p{1.8cm}|
                >{\centering\arraybackslash}p{2.0cm}}
\hline
\rowcolor{gray!20}
\textbf{Query} & \textbf{LLM Response} & \textbf{Sampled Responses} &
\textbf{Semantic N-gram} & \textbf{NLI Method} & \textbf{Prompt Method} \\
\hline

Who was Nikola Tesla, and what were his major contributions to science? &
Nikola Tesla was a Serbian-American inventor\ldots{} &
Tesla advanced AC \ldots{}; Tesla invented \ldots{}; Tesla's later life \ldots{} &
\textcolor{blue}{\textbf{0.33}} & \textcolor{blue}{\textbf{0.11}} & \textcolor{blue}{\textbf{0.09}} \\
\hline

Who was the first person to propose the theory of evolution, and what was their main idea? &
The first person to propose the theory of evolution was Charles Darwin\ldots{} &
Charles Darwin introduced \ldots{}; Lamarck proposed \ldots{}; Mendel's work focused \ldots{} &
\textcolor{black}{\textbf{0.71}} & \textcolor{black}{\textbf{0.89}} & \textcolor{black}{\textbf{0.93}}\\
\hline

\multicolumn{6}{l}{\textbf{Method Descriptions:}} \\
\multicolumn{6}{p{0.95\linewidth}}{%
\textbf{Semantic N-gram:} Evaluates responses using N-gram probabilities to compute hallucination scores and assess factuality.\newline
\textbf{NLI Method:} Sampled passages \( S_n \) merge with response \( r_i \), generating logits for entailment, contradiction, or neutral scores.\newline
\textbf{Prompt Method:} CoT prompting checks context support, classifying sentences as \textrm{``yes''} or \textrm{``no''} and computes hallucination scores.
} \\
\hline
\end{tabular}
\end{adjustbox}
\caption{Real-world hallucination detection on ChatGPT responses. Lower scores indicate higher confidence in factual accuracy.}
\label{tab:real_life_example}
\end{table*}

To assess \textcolor{black}{SelfCheck-Eval}'s behavior beyond controlled benchmarks, we conducted a preliminary evaluation on ChatGPT-generated responses for general knowledge queries. While limited in scope, this analysis provides initial insights into the framework's applicability to uncontrolled scenarios.

Table~\ref{tab:real_life_example} presents two illustrative cases with contrasting characteristics. The Tesla query represents a straightforward factual case where all three \textcolor{black}{methods} converge on low hallucination scores (0.09-0.33), indicating consistent content across responses. The evolution query demonstrates a different pattern: the main response attributes the proposal of evolutionary theory to Charles Darwin, while the sampled responses mention other figures such as Lamarck and Mendel. This divergence in content leads all \textcolor{black}{methods} to assign higher scores (0.71-0.93), indicating detected inconsistencies. These results demonstrate that the \textcolor{black}{methods} can provide consistent assessments across different types of content and respond to varying degrees of consensus among sampled responses. The convergence of all three \textcolor{black}{methods} in both cases suggests that the framework maintains its detection capabilities when applied to realistic scenarios outside controlled benchmarks. This preliminary exploration indicates promising directions for real-world deployment, while highlighting the need for more extensive validation across diverse query types and domains.

\section{Experimental Results}

\subsection{Results on WikiBio Dataset}

\begin{table*}[htbp]
\centering
\small
\renewcommand{\arraystretch}{0.70}
\begin{tabular}{l|l|l|l|c|c|c}
\toprule
\textbf{Method} & \textbf{LLM} & \textbf{Size} & \textbf{Prompt} & \textbf{NonFact} & \textbf{Factual} & \textbf{Ranking} \\
\midrule
Pre-trained & T5 small & 60M & ZS& 70.04  & 25.55 & -9.78 \\
Pre-trained & GPT2 & 124M & ZS &71.67 & 29.26 & 0.68  \\
Pre-trained & Roberta Large & 355M & ZS & 76.44  & 31.20 & 6.07  \\
Pre-trained & Phi-3 & 3.8B & ZS & 89.27 & 56.26 & 61.88 \\
Pre-trained & Llama 3.1 & 8B & ZS & 78.90 & 37.12 & 34.25 \\
Pre-trained & Mistral & 7B & ZS & 86.13 & 58.60 & 41.15 \\
Pre-trained & Gemma & 7B & ZS & 75.41 & 30.87 & 4.80 \\
Pre-trained &  T5 small & 60M & CoT& 69.97  & 24.79 &  -10.95  \\
Pre-trained &  GPT2  & 124M & CoT& 72.01 & 30.59 & 3.55   \\
Pre-trained &  Roberta Large& 355M & CoT&  73.86  & 31.47 & 13.02  \\
Pre-trained &  Phi-3 & 3.8B & CoT&  75.08 & 29.67 & 13.79  \\
Pre-trained &  Llama 3.1& 8B & CoT&   63.88 & 21.75 & -17.11\\
Pre-trained &  Mistral & 7B & CoT& 81.91 & 46.10 & 43.49   \\
Pre-trained &  Gemma & 7B & CoT& 70.97 &  25.99 & -7.62 \\

Fine-tuned & Phi-3 & 3.8B & - & 92.87 & 65.25 & 73.54 \\
Fine-tuned & Llama 3.1 & 8B & - & 76.85 & 29.71 & 8.91 \\
Fine-tuned & Mistral & 7B & - & 92.68 & 67.10 & 75.63 \\
Fine-tuned & Gemma & 7B & - & 83.47 & 43.12 & 50.98 \\
\bottomrule
\end{tabular}
\caption{Results for Pre-trained and Fine-tuned Methods on WikiBio Dataset.}
\end{table*}

\begin{table}[H]
\centering
\small
\renewcommand{\arraystretch}{0.70}
\begin{tabular}{l|l|l|l|c|c|c}
\toprule
\textbf{Method} & \textbf{LLM} & \textbf{Size} & \textbf{Prompt} & \textbf{NonFact} & \textbf{Factual} & \textbf{Ranking} \\
\midrule

Pre-trained & GPT2 & 124M & ZS &85.92 & 13.19 & -2.22  \\
Pre-trained & Phi-3 & 3.8B & ZS & 90.34 & 19.16 & 12.75 \\
Pre-trained & Llama 3.1 & 8B & ZS & 87.76 & 21.10 & 3.16 \\
Pre-trained & Mistral & 7B & ZS & 91.33 & 17.11 & 2.57 \\
Pre-trained & Gemma & 7B & ZS & 93.38 & 20.38 & 21.76 \\
Pre-trained &  GPT2  & 124M & CoT& 86.58 & 14.25 &  1.65 \\
Pre-trained &  Phi-3 & 3.8B & CoT& 87.76 & 21.10 &3.16   \\
Pre-trained &  Llama 3.1& 8B & CoT& 85.33 & 13.13 & -4.07    \\
Pre-trained &  Mistral & 7B & CoT& 91.33 &  17.11 & 2.57   \\
Pre-trained &  Gemma & 7B & CoT& 90.07 & 13.88 & 0.5 \\
Fine-tuned & Phi-3 & 3.8B & - & 93.38 & 20.38 & 21.76 \\
Fine-tuned & Llama 3.1 & 8B & - & 79.63 & 9.74 & -15.44\\
Fine-tuned & Mistral & 7B & - & 92.91 & 17.37 & 20.05 \\
Fine-tuned & Gemma & 7B & - & 82.58 & 10.70 & -16.77 \\
\bottomrule
\end{tabular}
\caption{Results for Pre-trained and Fine-tuned Methods on AIME Dataset.}
\end{table}

\end{document}